\documentclass[lettersize,journal]{IEEEtran}
\usepackage{amsmath,amsfonts}
\usepackage{algorithmic}
\usepackage{array}
\usepackage{subcaption}
\usepackage{textcomp}
\usepackage{stfloats}
\usepackage{url}
\usepackage{verbatim}
\usepackage{graphicx}
\usepackage{xcolor}
\usepackage{booktabs}
\usepackage{bigints}
\usepackage{placeins}

\newcommand{\x}{{\bf x}}

\newcommand{\blue}[1]{\textcolor[rgb]{0,0,0}{#1}}

\def\BibTeX{{\rm B\kern-.05em{\sc i\kern-.025em b}\kern-.08em
    T\kern-.1667em\lower.7ex\hbox{E}\kern-.125emX}}
\usepackage{balance}

\begin{document}
\title{Image Statistics Predict the Sensitivity of \\ Perceptual Quality Metrics}

\author{Alexander Hepburn\thanks{A. Hepburn and R. Santos-Rodriguez are with the School of Engineering Mathematics and Technology, University of Bristol, Bristol, United Kingdom}, Valero Laparra,
\thanks{V. Laparra and J. Malo are with the Image Processing Lab, Universitat de Valencia, Valencia, Spain} Raul Santos-Rodriguez, Jesus Malo
}

\markboth{}%
{How to Use the IEEEtran \LaTeX \ Templates}

\maketitle

\begin{abstract}
Previously, Barlow and Attneave hypothesised a link between biological vision and information maximisation. Following Shannon, information was defined using the probability of natural images. Several physiological and psychophysical phenomena have been derived from principles like info-max, efficient coding, or optimal denoising. However, it remains unclear how this link is expressed in mathematical terms from image probability. Classical derivations were subjected to strong assumptions on the probability models and on the behaviour of the sensors. Moreover, the direct evaluation of the hypothesis was limited by the inability of classical image models to deliver accurate estimates of the probability. Here, we directly evaluate image probabilities using a generative model for natural images, and analyse how probability-related factors can be combined to predict the sensitivity of state-of-the-art subjective image quality metrics, a proxy for human perception. We use information theory and regression analysis to find a simple model that when combining just two probability-related factors achieves 0.77 correlation with subjective metrics. This probability-based model is validated in two ways: through direct comparison with the opinion of real observers in a subjective quality experiment, and by reproducing basic trends of classical psychophysical facts such as the Contrast Sensitivity Function, the Weber-law, and contrast masking.
\end{abstract}


\section{Introduction}
One long standing discussion in artificial and human vision is about the principles that should drive sensory systems.
One example is Marr and Poggio's functional descriptions at the (more abstract) \emph{Computational Level} of vision~\cite{Marr76}. Another is Barlow's \emph{Efficient Coding Hypothesis}~\cite{At54,Barlow}, which suggests that vision is just an optimal information-processing task. In modern machine learning terms, the classical optimal coding idea qualitatively links the probability density function (PDF) of the images with the behaviour of the sensors.

An indirect research direction explored this link by proposing design principles for the system (such as infomax) to find optimal transforms and compare the features learned by these transforms and the ones found in the visual system, e.g. receptive fields~\cite{Olshausen1996,Bell97,hyvarinen2009natural} and nonlinear responses~\cite{simoncelli2001natural,Schwartz01}. This indirect approach, which was popular in the past due to limited computational resources, relies on gross approximations of the image PDF and on strong assumptions about the behaviour of the system. The set of PDF-related factors (or surrogates of the PDF) that were proposed as explanations of the behaviour in the above indirect approach is reviewed below in Sec.~\ref{sec:probvis}.

In contrast, here we propose a direct relation between the behaviour (sensitivity) of the system and the PDF of natural images. Following the preliminary suggestions in~\cite{hepburn2022on} about this direct relation, we rely on two elements. On the one hand, we use recent \emph{generative models} that represent large image datasets better than previous PDF approximations and provide us with an accurate estimate of the probability at query points~\cite{Oord2016pixelcnn,salimans2017pixelcnn++}. Whilst these statistical models are not analytical, they allow for sampling and log-likelihood prediction. They also allow us to compute gradients of the probability, which, as reviewed below, have been suggested to be related to sensible vision goals.

On the other hand, recent measures of \emph{perceptual distance} between images have recreated human opinion of subjective distortion to a great accuracy~\cite{laparra2016,zhang2018unreasonable,Hepburn2020perceptnet,ding2020image}. \blue{Whilst being just approximations to human sensitivity}, the sensitivity of the perceptual distances is a convenient computational description of the main trends of human vision that should be explained from scene statistics. 
\blue{These convenient proxies are important because (1) testing relations between the actual sensitivity and probability in \emph{many} points and directions of the image space is not experimentally feasible, and (2) once an approximate relation has been established %with approximate (metric) sensitivities,
the metric can be used to identify discriminative points and directions where one should measure actual (human) sensitivities.}

In this work, we identify the more relevant probabilistic factors that may be behind the non-Euclidean behaviour of perceptual distances and propose a simple expression to predict the perceptual sensitivity from these factors. First, we empirically show the relationship between the sensitivity of the metrics and different functions of the probability using conditional histograms. Then, we analyse this relationship quantitatively using mutual information, factor-wise and considering groups of factors. After, we use different regression models to identify a hierarchy in the factors that allow us to propose analytic relationships for predicting perceptual sensitivity. We perform an ablation analysis over the most simple closed-form expressions and select some solutions to predict the sensitivity given the selected functions of probability. The models are then experimentally validated, evaluating how well the models recreate noise visibility experiments. Finally, we evaluate these models with behaviour observed in the human visual system, and compare with the perceptual distance on their ability to replicate psychophysical experiments measuring human responses to contrast, frequency and luminance.

\section{Background and Proposed Methods}
In this section, we first recall the description of visual behaviour: the \emph{sensitivity of the perceptual distance}~\cite{hepburn2022on}, which is the feature to be explained by functions of the probability. Then we review the indirect probabilistic factors that were proposed in the past to explain visual perception. 

Finally, we introduce the tools to compute both behaviour and statistical factors: (i) the perceptual distances, (ii) the probability models, and (iii) how variations in the image space (distorted images) are chosen.  

\subsection{The problem: Perceptual Sensitivity}
Given an original image $\x$ and a distorted version $\tilde{\x}$, full-reference \emph{perceptual distances} are models, $D_p(\x, \tilde{\x})$, that accurately mimic the human opinion about the subjective difference between them.

In general, this perceived distance, highly depends on the particular image analysed and on the direction of the distortion. These dependencies make $D_p$ distinctly different from Euclidean metrics like the Root Mean Square Error (RMSE), $||\x - \tilde{\x}||_2$, which does not correlate well with human perception~\cite{Wang2009}. This characteristic dependence of $D_p$ is captured by the \emph{directional sensitivity of the perceptual distance}~\cite{hepburn2022on}:
\begin{equation}
    S(\x,\tilde{\x}) = \frac{D_p(\x, \tilde{\x})}{||\x-\tilde{\x}||_2}
    \label{eq:sensitivity}
\end{equation}
This is actually the numerical version of the directional derivative for a given $D_p$, and we will refer to it as the \emph{perceptual sensitivity} throughout the work.

In this approximation of human sensitivity using metrics, the increase of $D_p$ as $\tilde{\x}$ departs from $\x$ is conceptually equivalent to the empirical transduction functions that can be 
psychophysically measured in humans (e.g. by using maximum likelihood difference scaling~\cite{Maloney08}) along the direction $\x-\tilde{\x}$.
While specific directions have special meaning, e.g. distortions that alter contrast or brightness, here we are interested in the general behaviour of the sensitivity of perceptual metrics for different images, $\x$.

\subsection{Previous Probabilistic Explanations of Vision}
\label{sec:probvis} 
There is a rich literature trying to link the properties of visual systems with the PDF of natural images.
Principles and PDF-related factors that have been proposed in the past include (Table~\ref{tab:principles}): 

\textbf{(i)~Infomax for \emph{regular images}}. Transmission is optimised by reducing the redundancy, as in principal component analysis~\cite{Hancock91,Buschbaum83}, independent components analysis~\cite{Olshausen1996,Bell97,hyvarinen2009natural}, and, more generally, in PDF equalisation~\cite{Laughlin81} or factorisation~\cite{Malo10}. \blue{Perceptual discrimination has been related to transduction functions
obtained from the cumulative density~\cite{Wei17}, as in PDF equalisation.}

\textbf{(ii)~Optimal representations for \emph{regular images} in noise}. Noise may occur either at the front-end sensors~\cite{Miyasawa61,Atick90,Atick92} (optimal denoising), or at the internal response~\cite{Lloyd57,Twer01} (optimal quantisation, or optimal nonlinearities to cope with noise). While in denoising the solutions depend on the derivative of the log-likelihood~\cite{Raphan11,Vincent11}, optimal quantisation is based on resource allocation according to the PDF after saturating non linearities~\cite{Twer01,McLeod03,Series09,Ganguli11}. Bit allocation and quantisation according to nonlinear transforms of the PDF has been used in perceptually optimised coding~\cite{Macq92,Malo00}.
In fact, both factors considered above (infomax and quantisation) have been unified in a single framework where the representation is driven by the PDF raised to certain exponent~\cite{malo2006v1,Laparra12,Laparra15}.

\textbf{(iii)~Focus on \emph{surprising images} (as opposed to regular images)}. This critically different factor (surprise as opposed to regularities) has been suggested as a factor driving sensitivity to color~\cite{Gegen09,Wichmann02color}, and in visual saliency~\cite{Bruce06}. In this case, the surprise is described by the inverse of the probability (as opposed to the probability) as in the core definition of information~\cite{Shannon48}.

\textbf{(iv)~Energy (first moment, mean or average, of the signal)}.
Energy is the obvious factor involved in sensing. 
In statistical terms, the first eigenvalue of the manifold of a class of images represents the average luminance of the scene. The consideration of the nonlinear brightness-from-luminance response is a fundamental law in visual psychophysics (the Weber-Fechner law~\cite{Weber1846,Fechner1860}). It has statistical explanations related to the cumulative density~\cite{Laughlin81,malo2006v1,Laparra12} and using empirical estimation of reflectance~\cite{PurvesLottoPNAS11}. Adaptivity of brightness curves~\cite{Wittle92} can only be described using sophisticated non-linear architectures~\cite{Brainard05,martinez2018derivatives,Bertalmio20}. 

\textbf{(v)~Structure (second moment, covariance or spectrum, of the signal}. Beyond the obvious \emph{energy}, vision is about understanding the \emph{spatial structure}. The simplest statistical description of the structure is the covariance of the signal. The (roughly) stationary invariance of natural images implies that the covariance can be diagonalized in Fourier like-basis, $\Sigma(\x) = B \cdot \lambda \cdot B^\top$~\cite{Clarke81}, and that the spectrum of eigenvalues in $\lambda$ represents the average Fourier spectrum of images. The magnitude of the sinusoidal components compared to the mean luminance is the concept of \emph{contrast}~\cite{Michelson27,Peli90}, which is central in human spatial vision. Contrast thresholds have a distinct bandwidth~\cite{Campbell68}, which has been related to the spectrum of natural images~\cite{Atick92,Gomez20,Li22}.

\textbf{(vi) Heavy-tailed marginal PDFs in transformed domains} were used by classical generative models of natural images in the 90's and 00's~\cite{Simoncelli97,Malo00,hyvarinen2009natural,Oord14_t_student} and then these marginal models were combined through mixing matrices (either PCA, DCT, ICA or wavelets), conceptually represented by the matrix $B$ in the last column of Table~\ref{tab:principles}. In this context, the response of a visual system (according to PDF equalisation) should be related to non-linear saturation of the signal in the transform domain using the cumulative functions of the marginals. These explanations~\cite{Schwartz01,malo2006v1} have been given for the adaptive nonlinearities that happen in the wavelet-like representation in the visual cortex~\cite{Heeger92,Carandini12}, and also to define perceptual metrics~\cite{Daly90,Teo94,Malo97,Laparra10,laparra2016}.

\begin{table*}
\setlength\tabcolsep{2pt}
\centering
\caption{Probabilistic factors that have been proposed to predict sensitivity. A general problem of the classical literature (in Sec.~\ref{sec:probvis}) is that direct estimation of the probability at arbitrary images $p(\mathbf{x})$ was not possible. \blue{$\mu(\x)$ specifies the mean of $\x$, $\Sigma(\x)$ is the covariance of $\x$ and $B$ is a mixing matrix.}}\vspace{-0.2cm}
\label{tab:principles}
\begin{tabular}{|c|cc|c|c|c|c|}
\hline
(i) &
  \multicolumn{2}{c|}{(ii)} &
  (iii) &
  (iv) &
  (v) &
  (vi) \\ \hline
\begin{tabular}[c]{@{}c@{}}Information\\ Transmission\end{tabular} &
  \multicolumn{1}{c|}{\begin{tabular}[c]{@{}c@{}}Internal Noise\\ Limited Resolution\end{tabular}} &
  \begin{tabular}[c]{@{}c@{}}Acquisition Noise\\ Denoising\end{tabular} &
  Surprise &
  \begin{tabular}[c]{@{}c@{}}Signal Average\\ First Eigenvalue\end{tabular} &
  \begin{tabular}[c]{@{}c@{}}Signal Spectrum\\ All Eigenvalues\end{tabular} &
  \begin{tabular}[c]{@{}c@{}}Marginal Laplacian\\ Marginal nonlinearity\end{tabular} \\ \hline
\begin{tabular}[c]{@{}c@{}}$p(\x)$\\[0.15cm] $\log(p(\x))$\end{tabular} &
  \multicolumn{1}{c|}{\begin{tabular}[c]{@{}c@{}}$p(\x)^{\frac{1}{3}}$\\[0.15cm] $\frac{1}{3}\log(p(\x))$\end{tabular}} &
  \begin{tabular}[c]{@{}c@{}}$\frac{\nabla_{\!\x} p(\x)}{p(\x)}$\\[0.15cm] $J(\x) = \nabla_{\!\x} \log(p(\x))$\end{tabular} &
  \begin{tabular}[c]{@{}c@{}}$p(\x)^{-1}$\\[0.15cm] $-\log(p(\x))$\end{tabular} &
  \begin{tabular}[c]{@{}c@{}}$\mu(\x)$\\[0.1cm] $\log(\mu(\x))$\end{tabular} &
  \begin{tabular}[c]{@{}c@{}}$\frac{1}{\mu(\x)}\Sigma(\x)$\\[0.15cm] $\frac{1}{\mu(\x)} \,\, B \cdot \lambda \cdot B^\top$\end{tabular} &
  \begin{tabular}[c]{@{}c@{}}$\frac{1}{\mu(\x)} \,\, B \cdot \log(\lambda) \cdot B^\top$\\[0.15cm] $\bigintsss_{\,\x}^{\mathbf{\hat{x}}} \log( p(\x') ) \, d\x'$\end{tabular} \\ \hline
\end{tabular}
\vspace{-0.5cm}
\end{table*}

\subsection{Our proposal}
\label{sec:factors}
Here we revisit the classical principles in Table~\ref{tab:principles} to propose a \emph{direct} prediction of the sensitivity of state-of-the-art perceptual distances from the image PDF. The originality consists on the \emph{direct} computation of $p(\mathbf{x})$ through current generative models as opposed to the indirect approaches taken in the past to check the Efficient Coding Hypothesis (when direct estimation of $p(\mathbf{x})$ was not possible). 

In our prediction of the perceptual sensitivity we consider the following PDF-related factors:
\begin{equation}
\begin{aligned}
     \log(p(\x)) \, \, , \, \, \log(p(\tilde{\x})) \,\, , \,\, ||J(\x)|| \,\, , \,\, ||J(\tilde{\x})&|| \,\, , 
    (\x-\tilde{\x})^\top\!\! \cdot \! J(\x) \,\, , \,\, \\
    \mu(\x) \,\, ,  \sigma(\x) \,\, , \,\, &\bigintssss_{\,\x}^{\mathbf{\tilde{x}}} \log( p(\x') )\, d\x' 
\end{aligned}
\end{equation}

where $\x$ is the image, $\tilde{\x}$ is the distorted image, $J(\x)$ is the (vector) gradient of the log-likelihood at $\x$. The standard deviation of the image, $\sigma(\x)$ (as a single element of the covariance $\Sigma(\x)$), together with  $\mu(\x)$ capture the concept of RMSE contrast~\cite{Peli90} (preferred over Michelson contrast~\cite{Michelson27} for natural stimuli), and the integral takes the log-likelihood that can be computed from the generative models and accumulates it along the direction of distortion, qualitatively following the idea of cumulative responses proposed in equalisation methods~\cite{Laughlin81,malo2006v1,Laparra12,Laparra15,Wei17}.

\subsection{Representative perceptual distances}
\label{sec:metrics}
The most successful perceptual distances can be classified in four big families:
\textbf{(i) Physiological-psychophysical architectures.} These include~\cite{Daly90,Watson93,Teo94,Malo97,Laparra10,Martinez19,Hepburn2020perceptnet} and in particular it includes NLPD~\cite{laparra2016}, which consists of a sensible filterbank of biologically meaningful receptive fields and the canonical Divisive Normalization used in neuroscience~\cite{Carandini12}.
\textbf{(ii) Descriptions of the image structure.} These include the popular SSIM~\cite{Wang2004}, its (improved) multiscale version MS-SSIM~\cite{wang2003multiscale}, and recent version using deep learning: DISTS~\cite{ding2020image}.
\textbf{(iii)~Information-theoretic measures.} These include measures of transmitted information such as VIF~\cite{Sheikh06,Malo21}, and recent measures based on enforcing informational continuity in frames of natural sequences, such as PIM~\cite{Bhardwaj2020}.
\textbf{(iv) Regression models:} generic deep architectures used for vision tasks retrained to reproduce human opinion on distortion as LPIPS~\cite{zhang2018unreasonable}.

Here we use representative examples of the four families: MS-SSIM, NLPD, DISTS, PIM, and LPIPS. %Table~\ref{tab:MOS} in Appendix~\ref{sec:performance_metrics}
The performance of these measures in reproducing human opinion can be found in the supplementary material. Fig.~\ref{fig:images} visually demonstrates the image-dependent sensitivity to noise, effectively captured by the representative $D_p$ measure NLPD, in contrast to the less effective Euclidean distance, RMSE.

\begin{figure*}[b!]
    \begin{center}
    \hspace{0cm}\includegraphics[width=1.0
\textwidth]{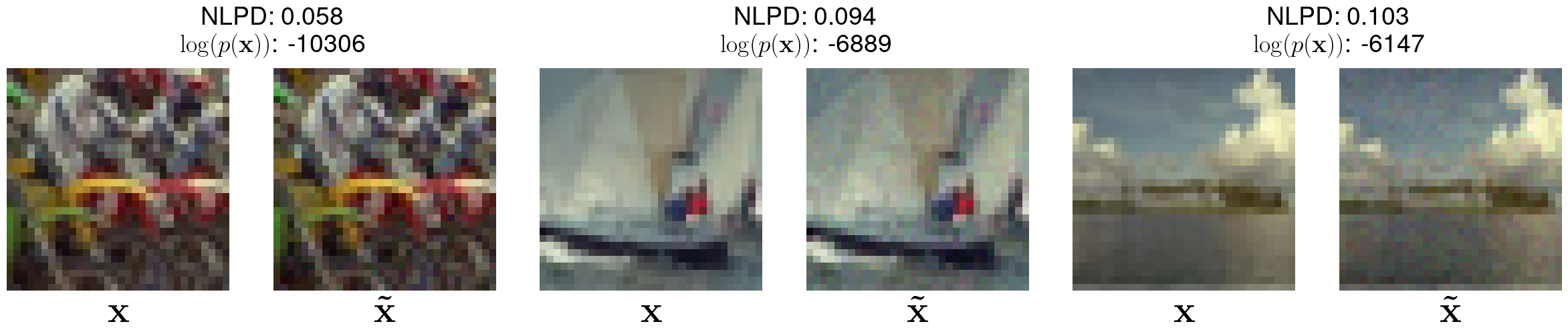}
\end{center}
    \vspace{-0.5cm}
    \caption{\textbf{The concept: visual sensitivity may be easy to predict from image probability.} Different images are corrupted by uniform noise of the same energy (on the surface of a sphere around $\x$ of Euclidean radius $\epsilon=1$), with the same RMSE = 0.018 (in [0, 1] range).
    Due to \emph{visual masking}~\cite{Legge80}, noise is more visible, i.e. human sensitivity is bigger, for smooth images.
    This is consistent with the reported NLPD distance, and interestingly,   
    sensitivity is also bigger for more probable images, $\log(p(\x))$ via PixelCNN++.
    }
    \label{fig:images}
\end{figure*}

\subsection{A convenient generative model}
\label{sec:est_prob}

Our proposal requires a probability model that is able to give an accurate estimate of $p(\mathbf{x})$ at arbitrary points (images) $\mathbf{x}$, so that we can study the sensitivity-probability relation at \emph{many} images. In this work we rely on PixelCNN++~\cite{salimans2017pixelcnn++}, since it is trained on CIFAR10~\cite{krizhevsky2009learning}, a dataset made up of small colour images of a \emph{wide range} of natural scenes. \blue{PixelCNN++ is an autoregressive model that fully factorises the PDF over all pixels in an image, where the conditional distributions are parameterised by a convolutional neural network}. PixelCNN++ also achieves the lowest negative log-likelihood of the test set, meaning the PDF is accurate in this dataset. This choice is convenient for consistency with previous works~\cite{hepburn2022on}, but note that it is not crucial for the argument made here. In this regard, recent interesting models~\cite{kingma2018glow,kadkhodaie23} could also be used as probability estimates.
% EXCUSE IF REVIEWERS ASK
%\footnote{\red{By the time we started this research, authors of \emph{Glow}~\cite{kingma2018glow} had released the weights only for human faces and bedrooms, which are not good representatives of generic natural images. Similarly, experiments in~\cite{kadkhodaie23} are focused on faces. Nevertheless, when trained in more general images this kind of models seem appropriate for our application.}}.

Fig.~\ref{fig:images} shows that cluttered images are successfully identified as less probable than smooth images by PixelCNN++, consistently with classical results on the average spectrum of natural images~\cite{Clarke81,Atick92,Malo00,simoncelli2001natural}.

\subsection{Distorted images}
\label{noisy_images}
There are two factors to balance when looking for distortions so that the definition of \emph{perceptual sensitivity} is meaningful. First; in order to understand the ratio in Eq.~\ref{eq:sensitivity} as a variation \emph{per unit of euclidean distortion}, we need the distorted image $\tilde{\x}$ to be close to the original $\x$. Secondly, the perceptual distances are optimised to recreate human judgements, so if the distortion is too small, such that humans are not able to see it, the distances are not trustworthy. 

As such, we propose to use additive Gaussian noise on the surface of a sphere of radius $\epsilon$ around $\x$. We choose $\epsilon$ in order to generate $\tilde{\x}$ where the noise is small but still visible for humans ($\epsilon=1$ in $32\times32\times3$ images with colour values in the range [0,1]). Python package IQM-Vis\footnote{\url{https://mattclifford1.github.io/IQM-Vis/}}was used to perform Just Noticeable Differences (JND) tests to determine small but visible noise.
Certain radius (or Euclidean distance) corresponds to a unique RMSE, with $\epsilon=1$ RMSE = 0.018 (about $2\%$ of the colour range).
Examples of the noisy images can be seen in Fig.~\ref{fig:images}. Note that in order to use the perceptual distances and PixelCNN++, the image values must be in a specific range. After adding noise we clip the signal to keep that range. Clipping may modify substantially the Euclidean distance (or RMSE) so we only keep the images where $\textrm{RMSE} \in [0.017, 0.018]$, resulting in 48,046 valid images out of 50,000. For each image, we sample 10 noisy images $\tilde{\x}$. Probabilities estimated using PixelCNN++ around this amount of noise are smooth as seen the supplementary material. %in Appendix~\ref{sec:smoothness}.

\section{Experiments}
Firstly, we empirically show the relation between the sensitivity and the candidate probabilistic factors using conditional histograms, and use information theory measures to tell which factors are most related to the sensitivity. Then, we explore polynomial combinations of several of these factors using regression models. Lastly, we restrict ourselves to the two most important factors and identify simple functional forms to predict perceptual sensitivity.

\subsection{Highlighting the relation: Conditional Histograms}
\label{sec:cond_hist}
\begin{figure*}[t]
    \centering
    \includegraphics[width=0.95\textwidth]{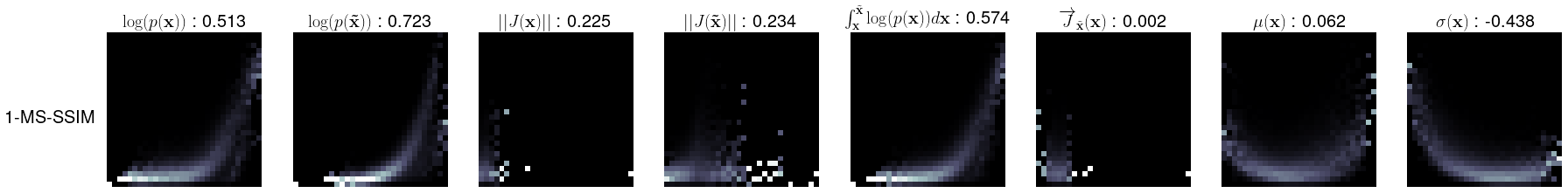}
    \includegraphics[width=0.95\textwidth]{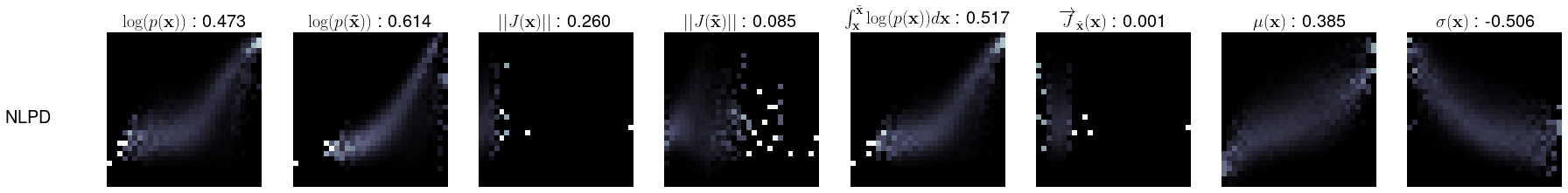}
    \includegraphics[width=0.95\textwidth]{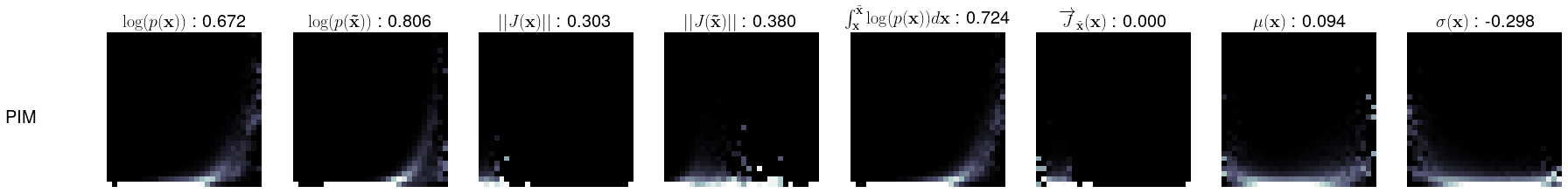}
    \includegraphics[width=0.95\textwidth]{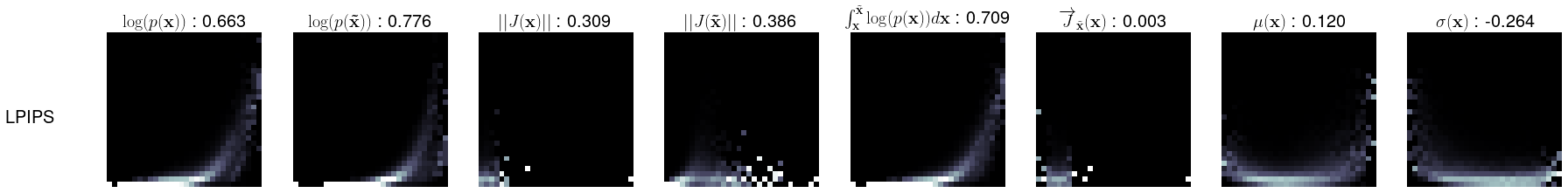}
    \includegraphics[width=0.95\textwidth]{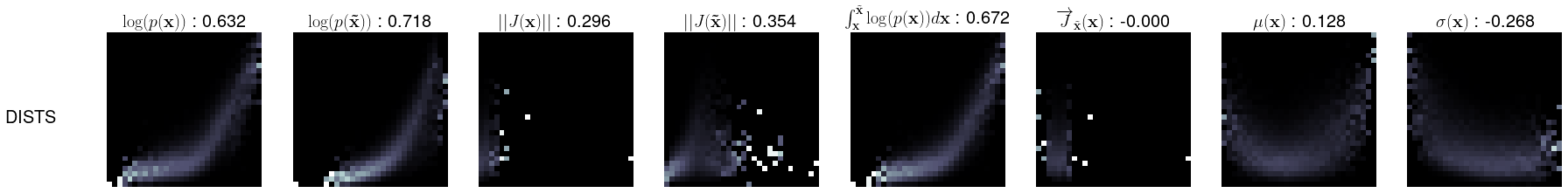}
    \includegraphics[width=.75\textwidth]{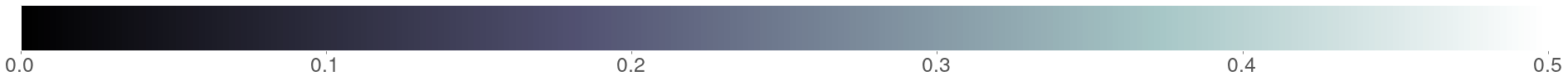}
    \caption{\footnotesize{Conditional histograms for the dataset described in Sec.\ref{noisy_images}. x-axis is the probability factor and y-axis is the sensitivity (Eq.~\ref{eq:sensitivity}) of the metric per row. Spearman correlation for each combination is in the title.}}
    \label{fig:cond_hist}
\end{figure*}

\blue{Conditional histograms in Fig.~\ref{fig:cond_hist} illustrate the relations between the sensitivity of each metric (Eq~.\ref{eq:sensitivity}) conditioned to different probabilistic factors using the dataset described in Sec.\ref{noisy_images}.} In this case, the histograms describe the probabilities $\mathcal{P}(S\in [b_{j-1}, b_j] | X=x)$ where $S$ is certain perceptual sensitivity partitioned into $m=30$ bins $[b_{j-1}, b_j)$, and $X$ is one of the possible probabilistic factors. This allows us to visually inspect which of the probabilistic factors are important for predicting sensitivity. The probabilistic factor used is given in each subplot title alongside with the Spearman correlation between perceptual sensitivity and the factor. 

For all perceptual distances, $\log(p(\x))$ has a high correlation, and mostly follows a similar conditional distribution. NLPD is the only distance that significantly differs, with more sensitivity in mid-probability images. We also see a consistent increase in correlation and conditional means when looking at $\log(p(\tilde{\x}))$, meaning the log-likelihood of the noisy sample is more indicative of perceptual sensitivity for all tested distances. Also note that the standard deviation $\sigma(\x)$ also has a strong (negative) correlation across the traditional distances, falling slightly with the deep learning based approaches. This is likely due to the standard deviation being closely related to the contrast of an image~\cite{Peli90}, and the masking effect: the sensitivity in known to decrease for higher contrasts~\cite{Legge80,Martinez19}. 
\blue{Note that, as anticipated above, in Eq.~\ref{eq:sensitivity}, the integration of the sensitivity along a direction is related to the transduction of the visual system along that specific direction.
With this in mind, the trends in the 2nd column or the last column in Fig.~\ref{fig:cond_hist} are consistent with high slope in the transduction function for low contrast images (of high probability).}
Note, also that measures that take into account both points, $\int^{\tilde{\x}}_\x \log(p(\x))d\x$ and $\overrightarrow{J}_{\tilde{x}}(\x)$, have lower correlation than just taking into account the noisy sample, with the latter being insignificant in predicting perceptual sensitivity.

\subsection{Quantifying the relation: Mutual Information}
\label{sec:MI}
To quantify the ability of probabilistic factors to predict perceptual sensitivity, we use information theoretic measures. Firstly, we use mutual information which avoids the definition of a particular functional model and functional forms of the features. This analysis will give us insights into which of the factors derived from the statistics of the data can be useful in order to later define a functional model that relates statistics and perception.

The mutual information has been computed using all 48046 samples and using the Rotation Based Iterative Gaussianisation (RBIG)~\cite{laparra2011iterative} as detailed here \cite{laparra2020information}. Instead of the mutual information value ($I(X, Y)$), we report the Information coefficient of correlation~\cite{Linfoot57} (ICC) since the interpretation is similar to the Pearson coefficient and allows for easy comparison, where $\textrm{ICC}(X, Y)= \sqrt{1- e^{-2 I(X, Y)}}$.

\begin{figure}[htb]
    \centering
    \includegraphics[width=0.9\columnwidth]{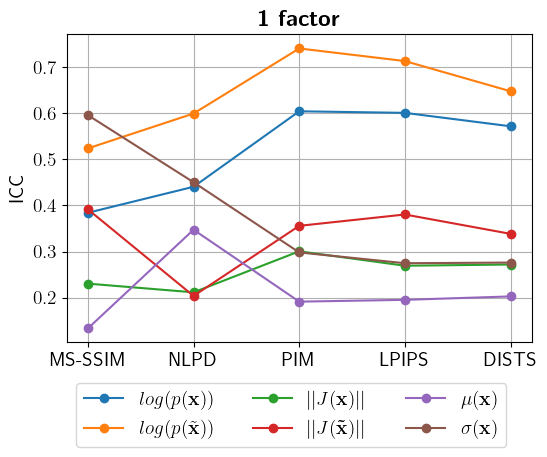} \\
    \includegraphics[width=1.0\columnwidth]{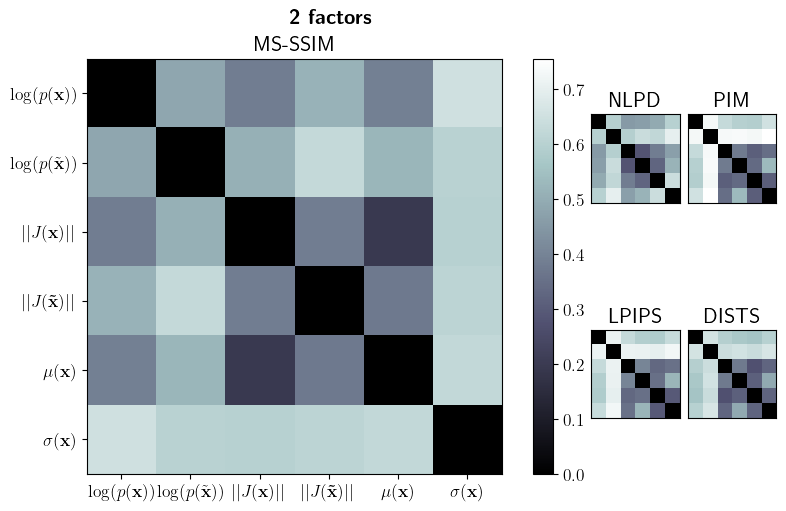}
    \caption{Information Coefficient of Correlation (ICC) between the sensitivity of different perceptual distances and isolated probabilistic factors (left: 1 factor) and different pairs of probabilistic factors (right: 2 factors). The considered factors are explicitly listed at the axes of the matrix of pairs for the MS-SSIM. The meaning of the entries for the matrices of the other distances is the same and all share the same colourbar for ICC.}
      \label{fig:MI_all}
\end{figure}

Fig.~\ref{fig:MI_all} shows the ICC between each isolated probabilistic factor and the sensitivity of different metrics. It is clear that the most important factor to take into account in most models (second in MS-SSIM) is the probability of the noisy image $\log(p(\tilde{\x}))$, a consistent result with the conditional histograms.
Once we select the $\log(p(\tilde{\x}))$ as the most important factor, we explore which other factor should be included as the second term. In order to do so we analyse the ICC between each possible combination of two factors with each metric (right panel of Fig.~\ref{fig:MI_all}). The matrix of possible pairs is depicted in detail for MS-SSIM (with factors in the axes and a colourbar for ICC), and corresponding matrices for the other metrics are also shown in smaller size. We can see that the general trend in four out of five metrics is the same: the pairs where $\log(p(\tilde{\x}))$ is involved have the maximum mutual information. The exception is MS-SSIM, which also behaves differently when using a single factor. In all the other cases the maximum relation with sensitivity is achieved when combining $\log(p(\tilde{\x}))$ with the standard deviation of the original image $\sigma(\x)$. 

The set of factors found to have a high relation with the variable to be predicted was expanded (one at a time) with the remaining factors. And we computed the mutual information and ICC for such expanded sets of factors. Reasoning as above (Fig.~\ref{fig:MI_all}, right), we found that the next in relevance was $\log(p(\x))$, then the mean $\mu(\x)$, then the gradient $||J(\x)||$, to finally take into account \emph{all} the probabilistic factors.   

Table~\ref{tab:MI} summarises the increase of ICC (the increase if information about the sensitivity) as we increase the number of probabilistic factors taken into account. In particular, we see that just by using $\log(p(\tilde{\x}))$ we can capture between $[0.55-0.71]$ of the information depending on the metric. Then, for all the metrics, the consideration of additional factors improves ICC, but this improvement saturates as new factors contain less independent information about the sensitivity.

\begin{table*}[b]
\setlength\tabcolsep{4pt}
\caption{Information Coefficient of Correlation between the sensitivity and groups of probabilistic factors.}
\centering
\begin{tabular}{lllllll}
\hline
\\ \textbf{Factors}                   & \textbf{MS-SSIM} & \textbf{NLPD} & \textbf{PIM} & \textbf{LPIPS} & \textbf{DISTS} & \textbf{mean} \\
1D: $\,\,\,\log(p(\tilde{x}))$  & 0.55 & 0.57 & 0.71 & 0.68 & 0.61 & 0.62\\
2D: \{$\log(p(\tilde{x}))$, $\sigma(x)$\}  & 0.57 & 0.66 & 0.72 & 0.69 & 0.63 & 0.65\\
3D: \{$\log(p(\tilde{x}))$, $\sigma(x)$, $\log(p(\x))$\}  & 0.68 & 0.68 & 0.72 & 0.69 & 0.65 & 0.68\\
4D: \{$\log(p(\tilde{x}))$, $\sigma(x)$, $\log(p(\x))$, $\mu(x)\}$ & 0.68 & 0.76 & 0.75 & 0.71 & 0.66  & 0.71 \\
5D: \{$\log(p(\tilde{\x}))$, $\sigma(x)$, $\log(p(\x))$, $\mu(x)$, $||J(x)||$\} & 0.68 & 0.78 & 0.75 & 0.73 & 0.66  & 0.72  \\
6D: all factors & 0.71 & 0.79 & 0.76 & 0.73 & 0.66  & 0.73 \\\hline   
\end{tabular}
\label{tab:MI}
\end{table*}

\subsection{Accurate but non-interpretable model: non-parametric regression} \label{sec:ML_regres}
The goal is to achieve an interpretable straightforward parametric function predicting sensitivity based on probabilistic factors. Non-parametric models, while less interpretable, provide a performance reference (upper bound) for assessing the quality of our developed interpretable models. To do so, we consider a random forest regressor~\cite{breiman2001random} to predict the perceptual sensitivity from the probabilistic factors and 2nd order polynomial combinations of the factors, including the inverse probability for both original and noisy images. In regression trees is easy to analyse the relevance of each feature and compare between models trained on different perceptual sensitivities. We use a held out test set of 30\% dataset in order to calculate correlations between predicted and ground truth. The average Pearson correlation obtained is 0.85, which serves as an illustrative upper bound reference for the interpretable functions proposed below in Sec.~\ref{sec:func_form}. We also trained a simple 3 layer multilayer perceptron on the same data and also achieved a correlation of 0.85.

Feature importances for the 6 probabilistic factors, according to a random forrest regressor trained to predict the perceptual sensitivity of the distances, can be found in the supplementary material. 
%shows the 6 probabilistic factors with the larger importance across perceptual distances and their relative relevance for each distance. 
The Pearson (Spearman) correlations indicate how good the regression models are at predicting sensitivity. $\log(p(\tilde{\x}))$ is the most important factor, which agrees with the information theoretic analysis (Sec.~\ref{sec:MI}). It has been suggested that the derivative of the log-likelihood is important in learning representations~\cite{bengio2013} since modifications in the slope of the distribution imply label change and the score-matching objective makes use of the gradient. However, we find that this derivative has low mutual information with human sensitivity and low influence in the regression.

\subsection{Interpretable model: Simple Functional Forms}
\label{sec:func_form}
In order to get simple interpretable models of sensitivity, in this section we restrict ourselves to linear combinations of power functions of the probabilistic factors. We explore two situations: a single-factor model (1F), and a two-factor model (2F).
According to the previous results on the ICC between the factors and sensitivity, the 1F model has to be based on $\log(p(\tilde{\x}))$, and the 2F model has to be based on $\log(p(\tilde{\x}))$ and $\sigma(\x)$.

\textbf{1-factor model (1F).} In the literature, the use of $\log(p(\tilde{\x}))$ has been proposed using different exponents (see Table~\ref{tab:principles}). Therefore, first, we analysed which exponents get better results for a single component, i.e. $S(\x,\tilde{\x}) = w_0 + w_1~(\log(p(\tilde{\x})))^\gamma$. We found out that there is not a big difference between different exponents. 
Therefore we decided to explore the simplest solution: a regular polynomial (with natural numbers as exponents). We found that going beyond degree 2 does not substantially improve the correlation (detailed results are in the supplementary material) %Table~\ref{tab:1D} in Appendix~\ref{sec:parameter_selection}).
We also explore a special situation where factors with different fractional and non-fractional exponents are linearly combined. The best overall reproduction of sensitivity across all perceptual metrics (Pearson correlation 0.73) can be obtained with a simple polynomial of degree two: 
\begin{equation}
    S(\x,\tilde{\x}) = w_0 + w_1~\log(p(\tilde{\x})) + w_2~(\log(p(\tilde{\x})))^2, 
\label{eq:1_factor}
\end{equation}
where the values for the weights for each perceptual distance are in the supplementary material. %Appendix~\ref{sec:app_coefficients_functionalform} (Table~\ref{tab:1D_coefs}). 

\textbf{2-factor model (2F).} In order to restrict the (otherwise intractable) exploration, we focus on polynomials that include the simpler versions of these factors, i.e. $\{ \log(p(\tilde{\x})), (\log(p(\tilde{\x})))^2 \}$  and $\{ \sigma(\x), \sigma(\x)^2, \sigma(\x)^{-1}\}$, and the simplest products and quotients using both.
We perform an ablation search of models that include these combinations as well as a LASSO regression~\cite{tibshirani1996regression} (details in the supplementary material).%Appendix~\ref{sec:parameter_selection} (Table~\ref{tab:app_2D})).
In conclusion, a model that combines good predictions and simplicity is the one that simply adds the $\sigma(\x)$ factor to the 1-factor model: 
\begin{equation}
    S(\x,\tilde{\x}) = w_0 + w_1~\log(p(\tilde{\x})) + w_2~(\log(p(\tilde{\x})))^2 + w_3~\sigma(\x)
\label{eq:2_factors}
\end{equation}
where the values of the weights for the different metrics are in the supplementary material. %Appendix~\ref{sec:app_coefficients_functionalform} (Table~\ref{tab:2D_coefs}).
The 2-factor model obtains an average Pearson correlation of $0.79$ across the metrics, which implies an increase of $0.06$ in correlation in regard to the 1-factor model. 
As a reference, a model that includes all nine analysed combinations for the two factors achieves an average $0.81$ correlation (not far from the performance obtained with the non-parametric regressor, $0.85$, in Sec.~\ref{sec:ML_regres}).

\textbf{Generic coefficients.} The specific coefficients for all the metrics are very similar. Therefore we normalise each set of coefficients with respect to $w_0$, and use the mean weights $\{w_0, w_1, w_2, w_3\}$ as a generic set of coefficients. It achieves very good correlation, 1F: $\rho = 0.74 \pm 0.02$; 2F: $\rho = 0.77 \pm 0.02$ (details are in the supplementary material) Throughout the rest of the paper, we will use 1F to denote the mean weights for the 1~factor model, and 2F for the 2~factor model.

% QUESTIONS
%
%  * In the 1-factor model did you propose a single expression with all possible coefficients (and then ablate)?... why not?
%  * In the 1-factor model, when you say d=3, you mean exponents 0, 1, 2, 3, right?
%  * The coefficients are undetermined up to a factor, right? why PIM is so off?
%
\section{Experimental Validation using natural images}
\label{sec:exp_val}
\begin{figure}[h]
    \centering
    \includegraphics[width=0.6\linewidth]{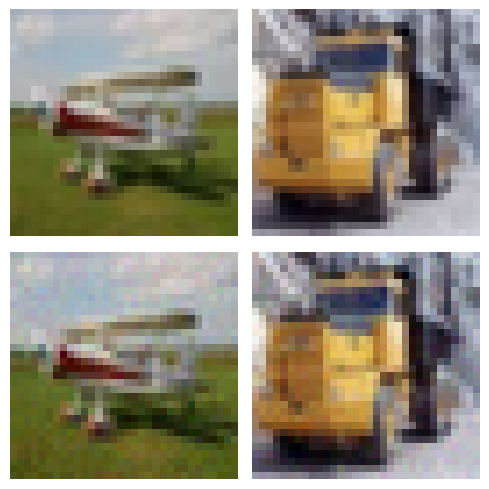}
    \caption{The method of quadruples~\cite{maloney2020measuring} experimental setup used, where participants are asked \emph{which difference is less visible; the left pair or right pair?}}
    \label{fig:noise_experiment_setup}
\end{figure}

In order to compare the 1F and 2F model responses when using the noisy images $\tilde{\x}$ to that of human observers, 
we conducted a psychophysical experiment using the method of quadruplets~\cite{maloney2020measuring}. In the experiment, participants are presented with two pairs of images, each containing an original $\x$ and a noisy $\tilde{\x}$, and are asked to select which reference-distorted pair is more similar. An example judgement can be found in Fig.~\ref{fig:noise_experiment_setup}. The winning pair is then given a point. For each participant, we gather $2$ judgements per comparison, using $30$ image pairs, resulting in $435$ total judgements per participant. We use Observer 1 as a base to compare against, and combine the other 5 participants' scores into a total score. This means that the scores come from a total of more than 2000 subjective judgements. The experiment was performed under the oversight of the Ethics Commission in Experimental Research of the Universitat de Valencia.

\begin{figure*}[htb]
    \centering
    \includegraphics[width=\textwidth]{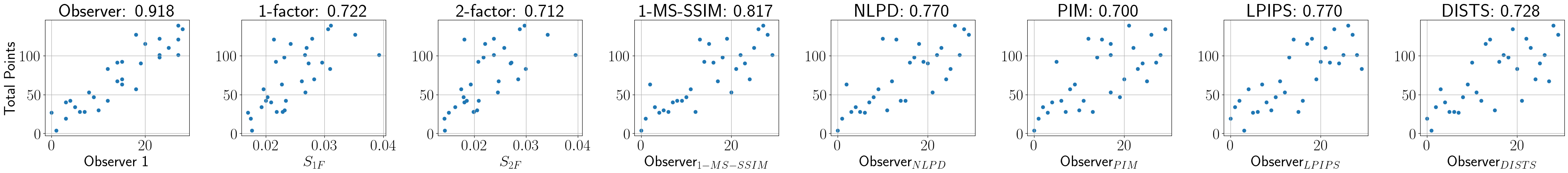}
    \caption{\textbf{Human Response to Noise}. Left most figure is Observer 1 vs the total points of the other 5 observers. For the 1F and 2F models, we plot the sensitivity $S$ of the distance against the total points of 5 observers. For the perceptual distances, we treat the distance as a participant and repeat the experiment. The Spearman correlation is in the title for each plot.}
    \label{fig:human_noise}
\end{figure*}

Then we compare the human results with the models. For the 1F and 2F model, we simply calculate the sensitivity (linear relationship with response since all images have the same RMSE). For the perceptual distances, we replicate the experiment but use the model as the participant. 

Fig.~\ref{fig:human_noise} shows the scores of Observer 1, our 1F and 2F models and various perceptual distances, compared to the total scores of 5 observers. The Spearman correlation is in the title of each plot.

The correlation of the sensitivity of an human observer with regard to the other observers is around 0.92. That stresses the fact that, although the test is hard since we are almost in the limit of visibility, humans respond not exactly the same but very similar. The performance of the perceptual metrics lies in between 0.7-0.8 of correlation. The models based just on image statistics (1F and 2F) have a correlation very similar to the metrics based on perception.

\section{Validation: reproduction of classical psychophysics using synthetic stimuli}
\label{validation}
Variations of human visual sensitivity depending on luminance, contrast, and spatial frequency were identified by classical psychophysics~\cite{Weber1846,Legge80,Campbell68,Georgeson75} way before the advent of the state-of-the-art perceptual distances considered here. Therefore, the reproduction of the trends of those classic behaviours is an independent way to validate the proposed models since it is not related to the recent image quality databases which are at the core of the perceptual distances.

We compare psychophysical results of our generic coefficient 1 and 2 factor models with the reference image quality metrics used throughout the paper; MS-SSIM, NLPD, PIM, LPIPS and DISTS.

In order to characterise the response of the models to different frequencies and contrast, we use 2D spatial Gabor functions~\cite{daugman1985uncertainty}, a type of linear filter that has shown to be present in visual systems of mammals~\cite{daugman1985uncertainty} and used to test models for human psychophysical behaviour~\cite{li2022contrast}.
Gabors are generated with different frequencies using the following definition;
% Definition of Gabor and Equation
\begin{equation}\label{eq:gabor}
    h(x, y) = \exp \Big[{-\frac{x^2 + y^2}{2 \sigma^2}}\Big] \cdot \sin(2\pi \omega x)
\end{equation}
where $h(x, y)$ is the value of the Gabor at coordinates $(x, y)$ in degrees, $\sigma$ is the width of the Gaussian window and $\omega$ is the frequency of the Gabor in cycles per degree. We have no rotation of the Gaussian envelope, and no rotation of the grating - it is strictly horizontal. Examples of Gabors generated in this manner can be seen in Fig~\ref{fig:csf_stimuli}. Throughout the experiments, Gabors are generated with width $\sigma=0.35$.

Additionally, the frequency response of humans changes with contrast, for which we wish to test the models. We define the contrast $C$ of an image as $C = \frac{\sqrt{2} \sigma}{\mu}$ where $\sigma^2$ is the standard deviation of the luminance, and $\mu$ is the mean luminance. We alter the $\mu$ and $\sigma^2$ of an image to achieve a certain contrast $C$. Under sinusoids, this contrast generalises to Michelson contrast~\cite{Michelson27}, whilst also being valid for Gabors.

The synthetic stimuli are generated in luminance, $L$ in $cd/m^2$, and must be transformed into digital values in order to evaluate models. Assuming a gamma correction with a power curve of $\x^2$, the digital values are given by $\x = \sqrt{\frac{L}{220}} * 255$
where $220 cd/m^2$ is the assumed maximum luminance of the monitor used to observe the images. This results in a digital image in the range $[0, 255]$. 

For some experiments, we wish to see the sensitivity of the models. The 1F and 2F models directly predict sensitivity, whereas for the perceptual distances we use the definition in Eq~\ref{eq:sensitivity}. To measure response, for the 1F and 2F models we predict the sensitivity, which is the derivative of the response~\cite{hepburn2022on}. Then we compute the response by integrating the derivative as in~\cite{Watson97}. For the perceptual distances, we take the distance between a test and a background, whether that background be a grating, as in the masking experiment, a flat image in the luminance experiment, or a flat gray image in the contrast sensitivity experiment.

\subsection{Weber Law}
\begin{figure}
    \centering
    \includegraphics[width=\linewidth]{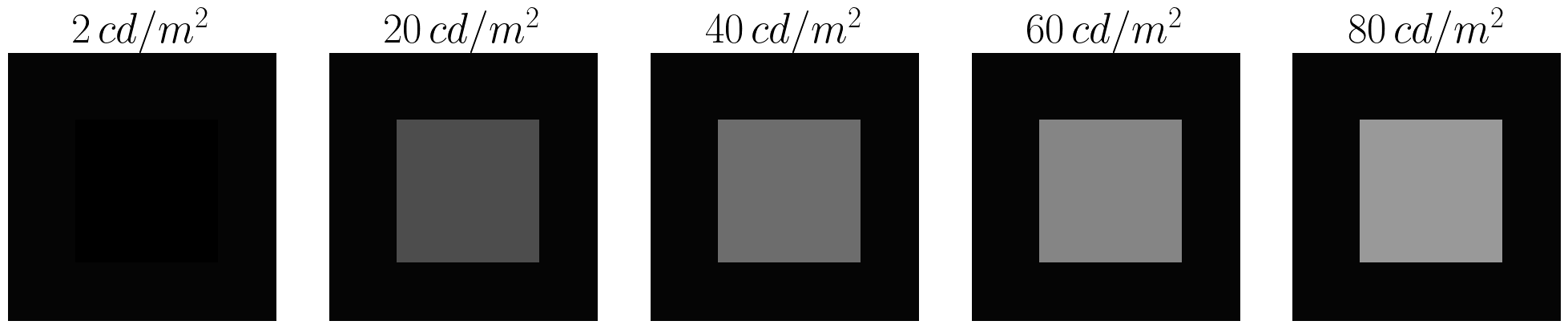}
    \caption{Stimuli used to predict model response to luminance.}
    \label{fig:weber_stimuli}
\end{figure}

% TODO: Show images with luminance patch
The first experiment is related to the known saturation of the brightness response (reduction of sensitivity) for higher luminances: the Weber law~\cite{Weber1846}. We use an experiment from experimental psychology~\cite{Wittle92}, consisting of a series of images with a flat background and a stimuli of varying luminance in the center. Subjects then adjusted the luminance of the stimuli, with the aim of creating equal-interval luminance stimuli, and humans response to luminance can be derived from this. We recreate this experiment, where the background luminance is fixed, and the luminance of the stimuli is increased. Examples of this are in Fig.~\ref{fig:weber_stimuli}. Previously the background luminance was changed to see the interaction with the luminance of the stimuli, creating a brightness scale~\cite{Fairchild05}, however we keep the background fixed to a flat black image.

We test stimuli with luminance in the range of $[2, 80]cd/m^2$, where $2 cd/m^2$ is the minimum value observable in a monitor and above $80 cd/m^2$ the response should be saturated. Examples of the stimuli can be found in Fig.~\ref{fig:weber_stimuli}. For the response of the perceptual distances, we simple take the distance between the flat background and the combined background and stimuli. For the 1F and 2F models, we compute the sensitivity then integrate to get the response.

\begin{figure*}[h]
    \centering
    \begin{subfigure}[t]{0.99\textwidth}
        \centering
        \includegraphics[width=\textwidth]{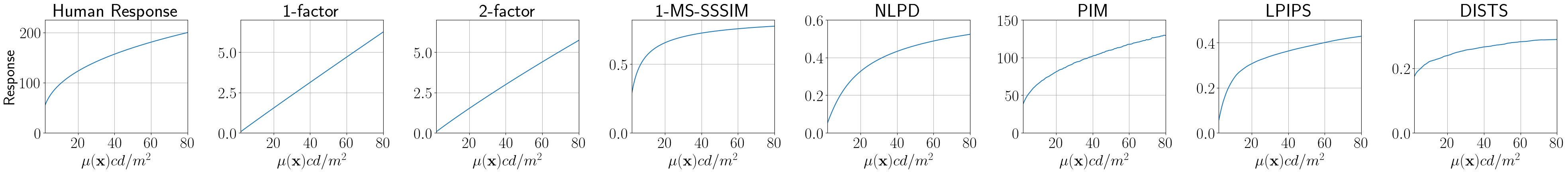}
        \caption{\textbf{Luminance (Weber Law)} Left most figure is the human response to different test luminances at a fixed background luminance~\cite{Stiles00}.The other plots are the various models we wish to test.}
        \label{fig:weber_law}
    \end{subfigure}
    \begin{subfigure}[t]{0.99\textwidth}
        \centering
        \includegraphics[width=\textwidth]{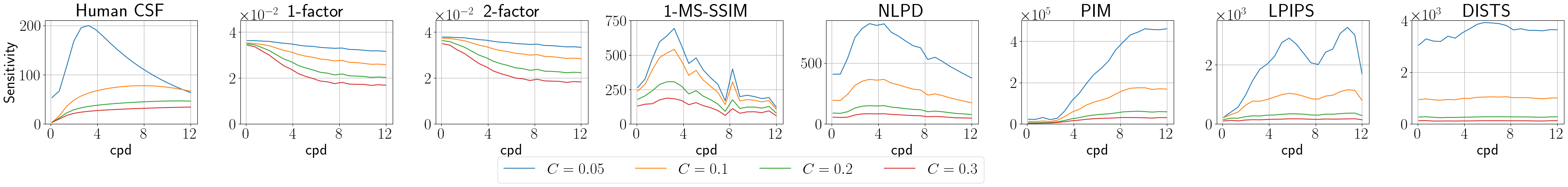}
        \caption{\textbf{Contrast Sensitivity Function (CSF)} Left most figure is the human sensitivity to different frequencies of Gabors at different contrasts~\cite{Campbell68,Watson02}. The other plots are the various models we wish to test.}
        \label{fig:csf_response}
    \end{subfigure}
    \begin{subfigure}[t]{0.99\textwidth}
        \centering
        \includegraphics[width=\textwidth]{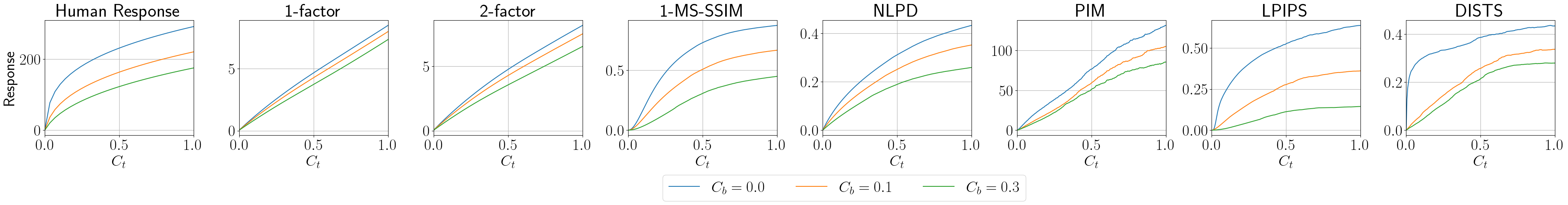}
        \caption{\textbf{Masking in contrast response}. Left most figure is the human response to different frequencies at different background contrasts $C_b$ over a range of test contrasts $C_t$ \cite{Watson97,Ross91,Foley94}. The other plots are the same but using various models.}
        \label{fig:masking_response}
    \end{subfigure}
    \caption{}
    \label{fig:psycho_tests}
\end{figure*}

Fig.~\ref{fig:weber_law} shows the responses of the models for different luminance stimuli. Humans exhibit a sharp increase in response at low luminance, and saturation of response at high luminance, as seen in the human response column. The perceptual distances all show a degree of both sharp increase and saturation, with NLPD and PIM achieving an almost identical response curve to humans. The 1F and 2F model are both linear in their response, maintaining low response in low luminance but not displaying saturation according to Weber law.

%We define contrast as $c = \sigma(\x)\sqrt{2}/ \mu(\x)$, which is applicable to natural images and reduces to Michelson contrast for gratings~\cite{Peli90}.
%We edited images from CIFAR10 to have a low contrast level ($0.1$) and we explored the range of luminances $[0, 70] cd/m^2$ so that the digital values still were in the [0,255] range.  We then averaged the sensitivities for 1000 images \blue{in CIFAR-10}. Averaged sensitivities computed using the \emph{generic coefficients} over 100 noise samples for both 

\subsection{Contrast Sensitivity Function}
%We compute the sensitivity to sinusoidal gratings of different frequencies but the same energy (or contrast).
The second experiment is related to the concept of Contrast Sensitivity Function (CSF)~\cite{Campbell68}: the human transfer function in the Fourier domain, which
decays and changes its shape with contrast as shown insuprathreshold contrast matching experiments~\cite{Georgeson75}. In this regard, we show the human CSF at contrasts $C=[0, 0.3]$ according to psychophysical models~\cite{Watson02,Malo97} using data from experiments involving threshold and suprathreshold visibility.

\begin{figure}[h]
    \centering
    \includegraphics[width=0.99\linewidth]{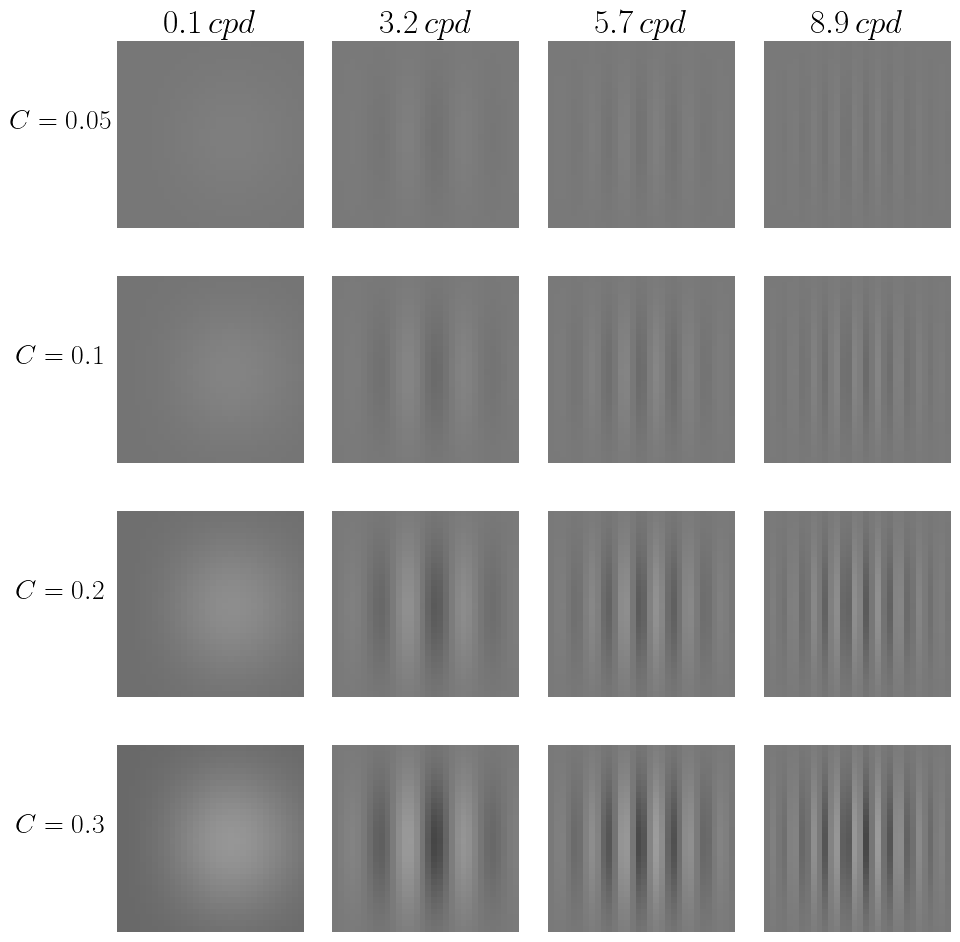}
    \caption{Example stimuli with different frequencies $\omega=\{0.1, 3.0, 5.9, 9.0\} cpd$(cycles per degree) and at varying contrast contrast $C=\{0.1, 0.2, 0.4, 0.6\}$.}
    \label{fig:csf_stimuli}
\end{figure}

As we cannot generates Gabors with contrast $0$, we approximate this to test the models using an extremely low contrast of $C=0.05$ in order to avoid the quantisation factor as the models work with 8-bit images. Gabors are generated with a contrasts of $C=\{0.05, 0.1, 0.2, 0.3\}$. We then evaluate the response of the models to Gabors of frequencies in the range $[1,12]$ cycles per degree (cpd). Examples of the stimuli are in Fig~\ref{fig:csf_stimuli}.

%Barely visible noise in $\epsilon=1$ sphere was added to the gratings to generate the distorted stimuli $\tilde{\x}$ in order to test the 1F and 2F models.

PixelCNN++ produces log-likelihoods such that the 1F and 2F models will never arrive to $0$ response. The minimum possible response arises from the least likely image according to PixelCNN++. As such, we rescale the response such that $0$ corresponds to the response given by the models to Gaussian noise with unit variance at the highest contrast ($C=0.3$), the minimum possible response of our 1F and 2F models.

For the 1F and 2F model, we simply predict the sensitivity given the Gabor. For the perceptual distances, we calculate the distance between a flat grey image with the same mean luminance as the Gabor ($50cd/m^2$) and the Gabor. The sensitivity is then the derivative of the response, given by Eq.~\ref{eq:sensitivity}. For the human sensitivity, we use the OSA Standard Spatial Observer~\cite{Watson02}, obtained from measuring thresholds of visibility in humans. The assumed contrast is the absolute threshold contrast, below 0.005. We approximate this very low contrast with 0.05, as if we were to use the absolute threshold contrast, there would be noticeable quantisation error in the digital images, or all values in the image would be $0$.

The human CSF at threshold shows a band-pass behavior: a sharp increase in response until a peaks at around 3 cpd, and after a slow decline. Both MS-SSIM and NLPD display this behaviour, however the response curve of NLPD is the most similar. PIM and LPIPS both peak at high frequencies, behaviour that is the opposite of humans and DISTS remains quite flat over all frequencies. The 1F and 2F models do not show the sharp increase, instead the peak is at 1 cpd, however the decrease after and saturation for high frequencies can both be observed in the human CSF.

In the probabilistic models (as well as in all the perceptual metrics) the frequency sensitivity is reduced for progressively higher contrast as in humans. However, the band-pass to all-pass change of shape in the function is only reproduced by NLPD.

\subsection{Masking in contrast responses}
Masking describes the reduction of response to test stimuli based off the background~\cite{Legge80}. Properties of the background such as contrast, frequency and orientation can either increase or decrease the visibility of test stimuli. 

\begin{figure}[h]
    \centering
    \includegraphics[width=.49\textwidth]{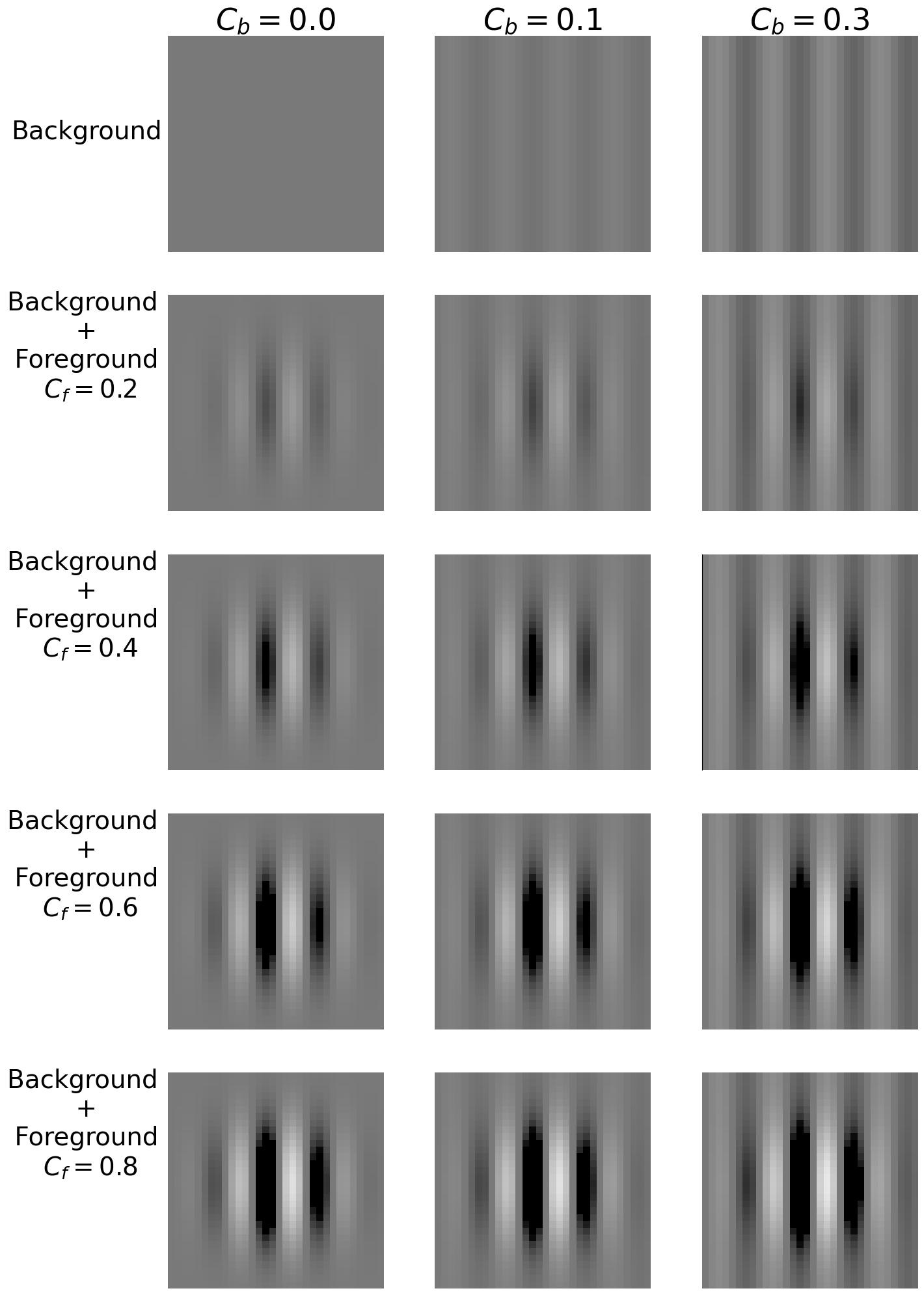}
    \caption{Stimuli used for masking experiment. We generate a background grating with $\omega = 4$ and contrast $C_b=\{0.0, 0.1, 0.3\}$ (top row), and increase the contrast of the test Gabor $\omega = 4$ in the range $C_t=[0, 1]$.}
    \label{fig:masking_stimuli}
\end{figure}

In this work, we wish to test the models for contrast masking, where a background of the same frequency and orientation as the stimuli is used. Both the contrast of the background and stimuli can be varied.
We generate sinusoidal gratings as backgrounds, with $\omega = 4$~cpd and contrasts $C_b=\{0.0, 0.1, 0.3\}$. The contrast of the Gabor test stimulus $\omega=4$ is then increased, in the range $[0, 1]$. Examples of the stimuli generated can be found in Fig.~\ref{fig:masking_stimuli}. The interaction between the background and test stimuli with the same wavelength should lower the visibility of the test, with a non-linear relationship between the background contrast $C_b$ and the response~\cite{Foley94,Watson97} as the contrast of the stimuli increases. % nonlinear in foreground contrast too

For 1F and 2F models, we calculate the sensitivity of the test and background combined, then calculate the cumulative sum in order to get the response from the sensitivity. For the perceptual distances, we simply take the distance between the background, and the combined background and test.

Fig.~\ref{fig:masking_response} shows the responses as we increase the contrast of the test $C_t$. In the human response, the effect of using a different value for the background contrast $C_b$ results in different response. Besides there is a sharp increase at low contrast of the test $C_t$, and saturation at higher contrasts. All perceptual distances achieve this behaviour to a varying degree of success. NLPD is the most similar, and strange behaviour can be observed in both PIM and DISTS. PIM is almost linear in the response to increased test contrast, whereas DISTS has an extremely sharp increase for $C_b=0.3$, extremely different behaviour than that observed at different background contrasts. The 1F and 2F also obtain a different behavior depending on the background contrast (more attenuation of the response with bigger contrast of the background) but responses are almost linear, with a slight non-linearity indicating a very minimal saturation. This is increased for the 2F model at $C_b=0.1$, indicating $\sigma(\x)$ in the 2F model increases the masking observed in these models.

% Regarding the sensitivity to gratings (left), results reproduce the decay of the CSF at high frequencies, its decay with contrast, and how the shape flattens with contrast, consistently with~\cite{Campbell68,Georgeson75}. For sensitivity at different luminances and contrasts (right) there is a simultaneous reduction in both, consistent with the literature~\cite{Weber1846,Legge80}. Interestingly, both models 1-F and 2-F follow the same trends.

% Justification for failure of model becuase probability model doesnt work for this kind of stimuli

% Put KS Tests - visually they are similar for masking although the KS Test says they're different (we have 50,000 samples to build the CDF with)

% \begin{figure}
%     \centering
%     \begin{subfigure}[t]{0.24\textwidth}
%         \centering
%         \includegraphics[width=
% \linewidth]{figs/masking_empirical.png}
%         \caption{Masking}
%         \label{fig:masking_emp}
%     \end{subfigure}
%     \begin{subfigure}[t]{0.24\textwidth}
%         \centering
%         \includegraphics[width=\linewidth]{figs/luminance_empirical.png}
%         \caption{Luminance}
%         \label{fig:lum_emp}
%     \end{subfigure}
%     \caption{Caption}
%     \label{fig:empirical}
% \end{figure}

\section{Discussion and Conclusion}
We show that several functions of image probability computed from a recent generative model share substantial information with 
the sensitivity of state-of-the-art perceptual metrics (mean ICC of 0.73, Sec.~\ref{sec:MI}). 
Alongside this, a Random Forest regressor that predicts sensitivity from polynomial combinations of these probabilistic factors obtains an average Pearson correlation of 0.85 with the sensitivity of the metrics (Sec.~\ref{sec:ML_regres}). 
According to the shared information, the factors can be ranked as: $\{\log(p(\tilde{\x})), \sigma(\x), \log(p(\x)), \mu(\x),  ||J(\x)||\}$, in agreement with the relevance given by the regression tree that uses the same features. 
These results are in contrast with intuitions about the importance of the distribution gradient~\cite{bengio2013}: we found that the gradient of the probability has low shared mutual information and low feature influence in the human sensitivity.

Using only $\log(p(\tilde{\x}))$ obtains an average ICC of 0.62, and a simple quadratic model including only this factor achieves an average of 0.73 Pearson correlation with perceptual metrics. 
%This  is due to simultaneously including information about the distribution of the original image $p(\x)$ and the specific direction of the distortion. The difference between $p(\x)$ and $p(\tilde{\x})$ is due to the direction moved to obtain the distorted image $\tilde{x}$. In the limit, with a small amount of Gaussian noise which is not visible to human observers, $p(\tilde{\x}) \approx p(\x)$).
%(in a differential context $p(\tilde{\x}) \approx p(\x)$), \red{and also about the specific direction of distortion}.  
It is important to note that, this factor includes two sources informations. On the one hand information about the distribution of the original image $p(\x)$, since in our experiments we test the differential behaviour, i.e. using a small amount of Gaussian noise, where $\tilde{\x} \approx \x$, and therefore $p(\tilde{\x}) \approx p(\x)$. On the other hand the difference between $p(\x)$ and $p(\tilde{\x})$ is due to the specific direction (noise realization) the original image $\x$ is moved to obtain the distorted image $\tilde{\x}$. Looking at Figure \ref{fig:MI_all}, the influence of the first effect (original image probability) is around $85\%$, and the second effect (specific direction of the distortion) is around $15\%$.

After an exhaustive ablation study keeping the factors with the highest ICC values, we propose simple functional forms of the sensitivity-probability relation using just 1 or 2 factors: Eqs.~\ref{eq:1_factor} (based on $\log(p(\tilde{\x}))$),
and Eq.~\ref{eq:2_factors} (based on $\log(p(\tilde{\x}))$ and $\sigma(\x)$)). These models obtain Pearson correlations with the metric sensitivities of $\rho = 0.74 \pm 0.02$; 2F: $\rho = 0.77 \pm 0.02$.

These simple functions of image probability are validated with a psychophysical experiment were we evaluate the sensitivity of humans for different natural images. Surprisingly, the proposed statistical models, which has no information about human behaviour, achieve a correlation similar to the psychophysical informed models. 
Additionally, we analysed the models in terms of the reproduction of classical human frequency sensitivity, the Weber law, and contrast masking (Sec.~\ref{validation}). 
In this case, the probabilistic models do reproduce the preference for low spatial frequencies. Moreover, the presence of a background masks (attenuates) the response to tests, also for different frequencies, and this attenuation increases with the energy of the background as in humans. However, the contrast responses and the luminance response do not display the saturation observed in human responses. While the agreement between the simple probabilistic models and humans in the experiment on subjective quality rating with \emph{natural images} is remarkable, the reproduction of classical psychophysics with \emph{synthetic stimuli} is not as good. 
However, note that state-of-the-art metrics also have problems to reproduce some of these curves: see the non-human nature of the frequency sensitivity of the deep-learning based techniques PIM, LPIPS and DISTS. 
This different behaviour with synthetic images versus natural images may be because the probabilistic model (trained for natural images) is not reliable with the synthetic stimuli. 
This is consistent with the fact that  perceptual metrics tuned to work with natural images may not reproduce visual effects illustrated with synthetic stimuli~\cite{Martinez19}.

\blue{This study inherits the limitations of the PixelCNN++ probability model, trained on a limited set of small images with restricted contrast, luminance, color, and content range. To ensure reliable $p(\x)$, we confined ourselves to a similar set of images. Additionally, we employ perceptual metrics as a surrogate for human perception. Despite correlations with humans, even state-of-the-art metrics have limitations. Consequently, further psychophysical work is essential to validate relationships with $p(\x)$. While the use of proxies is justified due to the impracticality of gathering sufficient psychophysical data, the insights gained can guide experimental efforts towards specific directions.}

Nevertheless, shared information and correlations are surprisingly high given that the statistics of the environment is not the only in driving factor in vision~\cite{Erichsen2012,Laughlin15}. In fact, this successful prediction of sensitivity from accurate measures of image probability with such simple functional forms is a renewed \emph{direct} evidence of the relevance of Barlow Hypothesis. 

\subsection{Acknowledgments}
\noindent Thanks to UKRI Turing AI Fellowship EP/V024817/1, MICIIN/FEDER/UE under Grants PID2020-118071GB-I00 and PDC2021-121522-C21, and by Generalitat Valenciana under Projects CIPROM/2021/056, CIAPOT/2021/9 and the computer resources at Artemisa, funded by the European Union ERDF and Comunitat Valenciana as well as the technical support provided by the Instituto de Física Corpuscular and IFIC (CSIC-UV).

\bibliographystyle{ieeetr}
\bibliography{ref}

\begin{IEEEbiographynophoto}{Alexander Hepburn}
received the M.Sc in 2017 and Ph.D. in 2022 funded by EPSRC doctoral training program (DTP), both from University of Bristol. Currently, he is a Senior Research Associate at University of Bristol as part of a Turing AI Fellowship in Interactive Annotations, working in imbalanced data, extending supervision beyond regular data-label pair and vision perception. 
% He received funding from the TAILOR project to visit the Institute of Optics at CSIC to develop methods of evaluating perceptual distances using data from psychophysical experiments.
His interests include vision perception, interpretable AI and fundamental machine learning.
\end{IEEEbiographynophoto}

%\begin{IEEEbiography}[{\includegraphics[width=1in,height=1.25in,clip,keepaspectratio]{photo_bio/valero_2022.png}}]
\begin{IEEEbiographynophoto}{Valero Laparra}
received the B.Sc degree in telecommunications engineering and the B.Sc. degree in the electronics engineering from the Universitat de Val\`encia, in 2005 and 2007, the B.Sc. degree in mathematics from the UNED in 2010, and the Ph.D. degree in computer science and mathematics from the Universitat de Val\`encia, in 2011. He is currently an Assistant Professor with the Escuela T\'ecnica Superior de Ingener\'ia, Universitat de Val\`encia. He is Researcher in the Image Processing Laboratory
\end{IEEEbiographynophoto}
%\end{IEEEbiography}

%\begin{IEEEbiography}[{\includegraphics[width=1in,height=1.25in,clip,keepaspectratio]{photo_bio/raul_4.jpg}}]
\begin{IEEEbiographynophoto}{Raul Santos-Rodriguez} received the Ph.D. degree in Telecommunication Engineering from Universidad Carlos III de Madrid, Spain in 2011. He is currently a UKRI Turing AI Fellow and Professor in Data Science and Intelligent Systems at the Department of Engineering Mathematics, University of Bristol. His main research interests focus on artificial intelligence and machine learning and their applications to music information retrieval, healthcare and remote sensing.
\end{IEEEbiographynophoto}
%\end{IEEEbiography}

%\begin{IEEEbiography}[{\includegraphics[width=1in,height=1.25in,clip,keepaspectratio]{photo_bio/jesus.JPG}}]
\begin{IEEEbiographynophoto}{Jesús Malo}
%(València, Spain, 1970)
received the M.Sc. and Ph.D. in Physics in 1995 and 1999 both from the Universitat de València. %He received the Vistakon European Research Award in 1994 for his work on physiological optics.
He worked on computational neuroscience during his postdoc stays at NASA Ames and NYU (2000–01) and as a visiting professor at Stanford and NYU (2013). 
%He was appointed Distinguished Benjamin Meaker Professor by the University of Bristol (2024).
He served as Editor of the IEEE Trans. Im. Proc., PLoS ONE, and Front. Neurosci, and has been on the committees of NeurIPS, ICLR, and the CIE. He was the first male member of the Asociación de Mujeres Investigadoras y Tecnólogas (AMIT). Currently, he is Professor of Vision Science and member of the Image and Signal Processing Group at the Universitat de València. His interests include %(but are not limited to)
models of low-level human vision, their relations with information theory, their applications to image processing and vision science experimentation.
%, and beauty in general.
\end{IEEEbiographynophoto}
%\end{IEEEbiography}

\pagebreak
\clearpage
\appendices

\section{Performance of the considered perceptual metrics}
\label{sec:performance_metrics}

Image quality metrics are able to predict visual human perception up to some level. In Table~\ref{tab:MOS} we show results of the different image quality metrics used in this work (see sec. \ref{sec:metrics}) when evaluated on human rates databases. Although the metrics are not perfect, it is clear that they are a good proxy of human perception. We show results on a traditional perceptual dataset using large images, TID2013~\cite{tid2013-data}, and a more recent dataset with smaller images (size $64\times64$) but more distortions as using neural networks based ones, such as super-resolution, BAPPS~\cite{zhang2018unreasonable}.

\begin{table}[h]
\centering
\footnotesize
\caption{Pearson and Spearman correlations with human opinion in TID2013, and agreement with human judgement (in \%) in BAPPS}.\vspace{-0.15cm}
\label{tab:MOS}
\begin{tabular}{|l|l|l|l|l|l|}
\hline
                                                                        & \textbf{MSSSIM}                                       & \textbf{NLPD}                                         & \textbf{PIM}                                          & \textbf{LPIPS}                                        & \textbf{DISTS}                                        \\ \hline
\begin{tabular}[c]{@{}l@{}}TID2013\\ $\rho_p$ ($\rho_s$)\end{tabular} & \begin{tabular}[c]{@{}l@{}}0.78\\ (0.80)\end{tabular} & \begin{tabular}[c]{@{}l@{}}0.84\\ (0.80)\end{tabular} & \begin{tabular}[c]{@{}l@{}}0.62\\ (0.65)\end{tabular} & \begin{tabular}[c]{@{}l@{}}0.74\\ (0.67)\end{tabular} & \begin{tabular}[c]{@{}l@{}}0.86\\ (0.83)\end{tabular} \\ \hline
\begin{tabular}[c]{@{}l@{}}BAPPS\\ (\%)\end{tabular}                    & 61.7                                                  & 61.5                                                  & 64.5                                                  & 69.2                                                  & 69.0                                                  \\ \hline
\end{tabular}
\end{table}

\section{Relevance of factors using random forest regressors}
\label{sec:app_coefficients_rndfrst}

\begin{figure*}[hb]
    \centering
    \includegraphics[width=.95\textwidth]{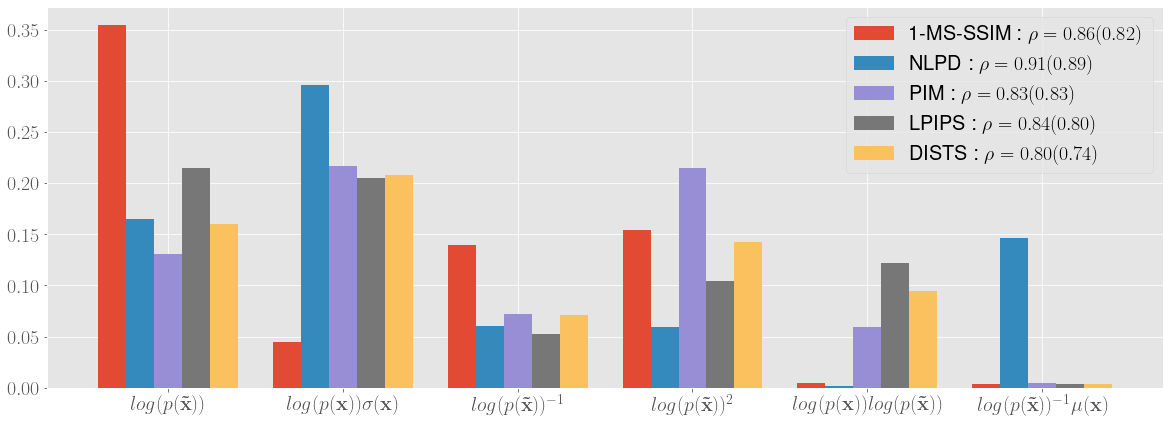}
    \caption{\footnotesize{Top 6 feature importance from Random Forest regressors trained on polynomial combinations of the probabilistic factors in order to predict perceptual sensitivity. A separate model was trained for each perceptual distance. In the legend we include the Pearson (Spearman) correlation between the predictions and ground truth for a held out test set 30\% of the dataset.}}
    \label{fig:regres_tree}
\end{figure*}

A random forest regression was fit using multiple combinations of the probability-related factors (see sec. \ref{sec:factors}) to predict the sensitivity of the different image quality metrics used in this work (see sec. \ref{sec:metrics}). A total of 55 combinations were introduced as inputs. Values of feature importance are normalised so that they sum to $1$ for each model for easy comparison. Fig.~\ref{fig:regres_tree} shows the most relevant ones selected by the random forest algorithm.

\section{Parameter selection for the functional forms}
\label{sec:parameter_selection}
Here we show the details for the selection of the parameters in Sec.~4. For the one-factor model, we tried different possibilities on the selected factor $\log(p(\tilde{\x}))$, details on the correlation of the different possibilities with sensitivity of perceptual measures are given in Table~\ref{tab:1D} and Sec~\ref{sec:func_form}.

\begin{table}[h]
\footnotesize
\caption{Pearson correlation obtained between the prediction of the model and the sensitivity for different IQMs. All models are designed using versions of $\log(p(\tilde{\x}))$ as input factor. The models are simple models with only \emph{one coefficient}, $S(\x,\tilde{\x}) = w_0 + w_1~(\log(p(\tilde{\x})))^\gamma$, or 
\emph{polynomials} of different degrees (d). The model \emph{Frac*} is a special polynomial with exponents: {[}$0.3,0.2,0.1,0,1,2,3${]}.}
\centering
    \label{tab:1D}
\begin{tabular}{lllllll}
\\         & \textbf{MSSIM} & \textbf{NLPD} & \textbf{PIM} & \textbf{LPIPS} & \textbf{DISTS} & \textbf{Mean} \\
\hline
\multicolumn{7}{c}{One coefficient}
\\
\hline\\
$\gamma$ = 1/10                       & 0.71           & 0.64          & 0.66         & 0.69           & 0.72           & \textbf{0.68} \\
$\gamma$ = 1/5                      & 0.71           & 0.64          & 0.66         & 0.69           & 0.72           & \textbf{0.68} \\
$\gamma$ = 1/3                       & 0.71           & 0.64          & 0.66         & 0.69           & 0.72           & \textbf{0.68} \\
$\gamma$ = 1/2                       & 0.71           & 0.63          & 0.66         & 0.69           & 0.72           & \textbf{0.68} \\
$\gamma$ = 2                       & 0.69           & 0.63          & 0.64         & 0.68           & 0.71           & \textbf{0.67} \\
$\gamma$ = -1                       & 0.72           & 0.65          & 0.66         & 0.71           & 0.73           & \textbf{0.69} 
\\
$\gamma$ = 1                       & 0.7            & 0.63          & 0.65         & 0.68           & 0.72           & \textbf{0.68} \\
\hline
\multicolumn{7}{c}{Polynomials}\\
\hline\\
d = 2                     & 0.76           & 0.65          & 0.75         & 0.76           & 0.74           & \textbf{0.73} \\
d = 3                     & 0.76           & 0.65          & 0.75         & 0.75           & 0.74           & \textbf{0.73} \\
d = 6                     & 0.75           & 0.64          & 0.73         & 0.74           & 0.74           & \textbf{0.72} \\
\emph{Frac*} & 0.76           & 0.65          & 0.76         & 0.76           & 0.74           & \textbf{0.73} \\
\hline
\end{tabular}
\end{table}

For the two-factors model, we took the ones obtained in Sec~4.1 (i.e. $b$ (bias), $\log(p(\tilde{\x}))$, and $(\log(p(\tilde{\x})))^2$), and factors of the standard deviation $\sigma_x$ as suggested by the mutual information in Sec.~\ref{sec:MI}. The combinations of the standard deviation have been alone: $\sigma_x$, $\frac{1}{\sigma_x}$,$\sigma_x^2$, and combined with the probability of the noisy image: $\frac{\log(p(\tilde{\x}))}{\sigma_x}$, $\frac{\sigma_x}{\log(p(\tilde{\x}))}$, and $\log(p(\tilde{\x})){\sigma_x}$. 

There are 9 candidates but we want the most compact and interpretable model. Analysing all the possible combinations is intractable so we are going to perform an ablation study: we are going to discard different candidates sequentially starting from the largest model (9 candidates). Besides, we are going to use LASSO regression with different amounts of regularisation to get models with different amounts of factors as comparison to the ablation study. Results in Table~\ref{tab:app_2D} are shown in descending number of factors used. For each step, we remove the factor (or factors) that less influence has in the correlation. Besides, we show the correlation given by LASSO models where the regularization parameter has been adjusted in order to have the same number of factors. A model with 6 factors (number 17) has the same correlation (0.81) as the one with all the factors (number 1). The best trade-off between the number of factors and correlation is for models 25, 26, and 27, with 4 factors and a correlation of 0.79. We chose as our final functional model in Sec.~4.2 the model 25 as its factors involve less computations.      

\begin{table*}[]
\centering
\setlength\tabcolsep{4pt}
\renewcommand{\arraystretch}{0.99}
\caption{Linear models using different combinations of $b$ (bias), $log(p(\tilde{x})) = p$ and $\sigma(x) = s$, where $1$ indicates the factor is included in the model, and $0$ it is ablated. Models with $*$ in the model number (\# M) have been computed using Lasso and a corresponding regularization parameter in order to have a particular number of combinations (\# C). For the perceptual metrics, the Pearson correlation between the predicted and ground truth on a test set of CIFAR10 is reported.}
\label{tab:app_2D}
\begin{tabular}{lllllllllllllllll}
\\ \hline
\multicolumn{1}{|l|}{\textbf{$b$}} & \multicolumn{1}{l|}{\textbf{$p$}} & \multicolumn{1}{l|}{\textbf{$p^2$}} & \multicolumn{1}{l|}{\textbf{$s$}} & \multicolumn{1}{l|}{\textbf{$s^2$}} & \multicolumn{1}{l|}{\textbf{$\frac{1}{s}$}} & \multicolumn{1}{l|}{\textbf{$\frac{p}{s}$}} & \multicolumn{1}{l|}{\textbf{$\frac{s}{p}$}} & \multicolumn{1}{l|}{\textbf{$p s$}} & \multicolumn{1}{l|}{\textbf{MSSIM}} & \multicolumn{1}{l|}{\textbf{NLPD}} & \multicolumn{1}{l|}{\textbf{PIM}} & \multicolumn{1}{l|}{\textbf{LPIPS}} & \multicolumn{1}{l|}{\textbf{DISTS}} & \multicolumn{1}{l|}{\textbf{Mean}}  & \multicolumn{1}{l|}{\textbf{\# C}} & \multicolumn{1}{l|}{\textbf{\#M}} \\ \hline
\multicolumn{1}{|l|}{1}          & \multicolumn{1}{l|}{1}          & \multicolumn{1}{l|}{1}                             & \multicolumn{1}{l|}{1}          & \multicolumn{1}{l|}{1}                             & \multicolumn{1}{l|}{1}            & \multicolumn{1}{l|}{1}            & \multicolumn{1}{l|}{1}            & \multicolumn{1}{l|}{1}            & \multicolumn{1}{l|}{0.85}           & \multicolumn{1}{l|}{0.78}          & \multicolumn{1}{l|}{0.81}         & \multicolumn{1}{l|}{0.81}           & \multicolumn{1}{l|}{0.78}           & \multicolumn{1}{l|}{\textbf{0.806}} & \multicolumn{1}{l|}{9}                & \multicolumn{1}{l|}{1}                \\ \hline
                                \multicolumn{17}{c}{\bf 8 combinations}             \\ \hline
\multicolumn{1}{|l|}{1}          & \multicolumn{1}{l|}{1}          & \multicolumn{1}{l|}{1}                             & \multicolumn{1}{l|}{0}          & \multicolumn{1}{l|}{1}                             & \multicolumn{1}{l|}{1}            & \multicolumn{1}{l|}{1}            & \multicolumn{1}{l|}{1}            & \multicolumn{1}{l|}{1}            & \multicolumn{1}{l|}{0.85}           & \multicolumn{1}{l|}{0.78}          & \multicolumn{1}{l|}{0.8}          & \multicolumn{1}{l|}{0.8}            & \multicolumn{1}{l|}{0.78}           & \multicolumn{1}{l|}{\textbf{0.802}} & \multicolumn{1}{l|}{8}                & \multicolumn{1}{l|}{2}                \\ \hline
\multicolumn{1}{|l|}{1}          & \multicolumn{1}{l|}{1}          & \multicolumn{1}{l|}{1}                             & \multicolumn{1}{l|}{1}          & \multicolumn{1}{l|}{0}                             & \multicolumn{1}{l|}{1}            & \multicolumn{1}{l|}{1}            & \multicolumn{1}{l|}{1}            & \multicolumn{1}{l|}{1}            & \multicolumn{1}{l|}{0.85}           & \multicolumn{1}{l|}{0.78}          & \multicolumn{1}{l|}{0.81}         & \multicolumn{1}{l|}{0.81}           & \multicolumn{1}{l|}{0.78}           & \multicolumn{1}{l|}{\textbf{0.806}} & \multicolumn{1}{l|}{8}                & \multicolumn{1}{l|}{3}                \\ \hline
\multicolumn{1}{|l|}{1}          & \multicolumn{1}{l|}{1}          & \multicolumn{1}{l|}{1}                             & \multicolumn{1}{l|}{1}          & \multicolumn{1}{l|}{1}                             & \multicolumn{1}{l|}{0}            & \multicolumn{1}{l|}{1}            & \multicolumn{1}{l|}{1}            & \multicolumn{1}{l|}{1}            & \multicolumn{1}{l|}{0.85}           & \multicolumn{1}{l|}{0.78}          & \multicolumn{1}{l|}{0.81}         & \multicolumn{1}{l|}{0.81}           & \multicolumn{1}{l|}{0.78}           & \multicolumn{1}{l|}{\textbf{0.806}} & \multicolumn{1}{l|}{8}                & \multicolumn{1}{l|}{4}                \\ \hline
\multicolumn{1}{|l|}{1}          & \multicolumn{1}{l|}{1}          & \multicolumn{1}{l|}{1}                             & \multicolumn{1}{l|}{1}          & \multicolumn{1}{l|}{1}                             & \multicolumn{1}{l|}{1}            & \multicolumn{1}{l|}{0}            & \multicolumn{1}{l|}{1}            & \multicolumn{1}{l|}{1}            & \multicolumn{1}{l|}{0.85}           & \multicolumn{1}{l|}{0.78}          & \multicolumn{1}{l|}{0.81}         & \multicolumn{1}{l|}{0.81}           & \multicolumn{1}{l|}{0.78}           & \multicolumn{1}{l|}{\textbf{0.806}} & \multicolumn{1}{l|}{8}                & \multicolumn{1}{l|}{5}                \\ \hline
\multicolumn{1}{|l|}{1}          & \multicolumn{1}{l|}{1}          & \multicolumn{1}{l|}{1}                             & \multicolumn{1}{l|}{1}          & \multicolumn{1}{l|}{1}                             & \multicolumn{1}{l|}{1}            & \multicolumn{1}{l|}{1}            & \multicolumn{1}{l|}{0}            & \multicolumn{1}{l|}{1}            & \multicolumn{1}{l|}{0.85}           & \multicolumn{1}{l|}{0.78}          & \multicolumn{1}{l|}{0.8}          & \multicolumn{1}{l|}{0.8}            & \multicolumn{1}{l|}{0.78}           & \multicolumn{1}{l|}{\textbf{0.802}} & \multicolumn{1}{l|}{8}                & \multicolumn{1}{l|}{6}                \\ \hline
\multicolumn{1}{|l|}{1}          & \multicolumn{1}{l|}{1}          & \multicolumn{1}{l|}{1}                             & \multicolumn{1}{l|}{1}          & \multicolumn{1}{l|}{1}                             & \multicolumn{1}{l|}{1}            & \multicolumn{1}{l|}{1}            & \multicolumn{1}{l|}{1}            & \multicolumn{1}{l|}{0}            & \multicolumn{1}{l|}{0.85}           & \multicolumn{1}{l|}{0.78}          & \multicolumn{1}{l|}{0.8}          & \multicolumn{1}{l|}{0.8}            & \multicolumn{1}{l|}{0.78}           & \multicolumn{1}{l|}{\textbf{0.802}} & \multicolumn{1}{l|}{8}                & \multicolumn{1}{l|}{7}                \\ \hline
                                \multicolumn{17}{c}{\bf 7 combinations}                                      \\ \hline
\multicolumn{1}{|l|}{0}          & \multicolumn{1}{l|}{1}          & \multicolumn{1}{l|}{1}                             & \multicolumn{1}{l|}{1}          & \multicolumn{1}{l|}{1}                             & \multicolumn{1}{l|}{1}            & \multicolumn{1}{l|}{1}            & \multicolumn{1}{l|}{1}            & \multicolumn{1}{l|}{0}            & \multicolumn{1}{l|}{0.83}           & \multicolumn{1}{l|}{0.75}          & \multicolumn{1}{l|}{0.77}         & \multicolumn{1}{l|}{0.77}           & \multicolumn{1}{l|}{0.77}           & \multicolumn{1}{l|}{\textbf{0.778}} & \multicolumn{1}{l|}{7}                & \multicolumn{1}{l|}{8}                \\ \hline
\multicolumn{1}{|l|}{1}          & \multicolumn{1}{l|}{0}          & \multicolumn{1}{l|}{1}                             & \multicolumn{1}{l|}{1}          & \multicolumn{1}{l|}{1}                             & \multicolumn{1}{l|}{1}            & \multicolumn{1}{l|}{1}            & \multicolumn{1}{l|}{1}            & \multicolumn{1}{l|}{0}            & \multicolumn{1}{l|}{0.82}           & \multicolumn{1}{l|}{0.78}          & \multicolumn{1}{l|}{0.77}         & \multicolumn{1}{l|}{0.78}           & \multicolumn{1}{l|}{0.77}           & \multicolumn{1}{l|}{\textbf{0.784}} & \multicolumn{1}{l|}{7}                & \multicolumn{1}{l|}{9}                \\ \hline
\multicolumn{1}{|l|}{1}          & \multicolumn{1}{l|}{1}          & \multicolumn{1}{l|}{0}                             & \multicolumn{1}{l|}{1}          & \multicolumn{1}{l|}{1}                             & \multicolumn{1}{l|}{1}            & \multicolumn{1}{l|}{1}            & \multicolumn{1}{l|}{1}            & \multicolumn{1}{l|}{0}            & \multicolumn{1}{l|}{0.82}           & \multicolumn{1}{l|}{0.78}          & \multicolumn{1}{l|}{0.76}         & \multicolumn{1}{l|}{0.77}           & \multicolumn{1}{l|}{0.76}           & \multicolumn{1}{l|}{\textbf{0.778}} & \multicolumn{1}{l|}{7}                & \multicolumn{1}{l|}{10}               \\ \hline
\multicolumn{1}{|l|}{1}          & \multicolumn{1}{l|}{1}          & \multicolumn{1}{l|}{1}                             & \multicolumn{1}{l|}{0}          & \multicolumn{1}{l|}{1}                             & \multicolumn{1}{l|}{1}            & \multicolumn{1}{l|}{1}            & \multicolumn{1}{l|}{1}            & \multicolumn{1}{l|}{0}            & \multicolumn{1}{l|}{0.84}           & \multicolumn{1}{l|}{0.78}          & \multicolumn{1}{l|}{0.8}          & \multicolumn{1}{l|}{0.8}            & \multicolumn{1}{l|}{0.78}           & \multicolumn{1}{l|}{\textbf{0.8}}   & \multicolumn{1}{l|}{7}                & \multicolumn{1}{l|}{11}               \\ \hline
\multicolumn{1}{|l|}{1}          & \multicolumn{1}{l|}{1}          & \multicolumn{1}{l|}{1}                             & \multicolumn{1}{l|}{1}          & \multicolumn{1}{l|}{0}                             & \multicolumn{1}{l|}{1}            & \multicolumn{1}{l|}{1}            & \multicolumn{1}{l|}{1}            & \multicolumn{1}{l|}{0}            & \multicolumn{1}{l|}{0.85}           & \multicolumn{1}{l|}{0.78}          & \multicolumn{1}{l|}{0.8}          & \multicolumn{1}{l|}{0.8}            & \multicolumn{1}{l|}{0.78}           & \multicolumn{1}{l|}{\textbf{0.802}} & \multicolumn{1}{l|}{7}                & \multicolumn{1}{l|}{12}               \\ \hline
\multicolumn{1}{|l|}{1}          & \multicolumn{1}{l|}{1}          & \multicolumn{1}{l|}{1}                             & \multicolumn{1}{l|}{1}          & \multicolumn{1}{l|}{1}                             & \multicolumn{1}{l|}{0}            & \multicolumn{1}{l|}{1}            & \multicolumn{1}{l|}{1}            & \multicolumn{1}{l|}{0}            & \multicolumn{1}{l|}{0.85}           & \multicolumn{1}{l|}{0.78}          & \multicolumn{1}{l|}{0.8}          & \multicolumn{1}{l|}{0.8}            & \multicolumn{1}{l|}{0.78}           & \multicolumn{1}{l|}{\textbf{0.802}} & \multicolumn{1}{l|}{7}                & \multicolumn{1}{l|}{13}               \\ \hline
\multicolumn{1}{|l|}{1}          & \multicolumn{1}{l|}{1}          & \multicolumn{1}{l|}{1}                             & \multicolumn{1}{l|}{1}          & \multicolumn{1}{l|}{1}                             & \multicolumn{1}{l|}{1}            & \multicolumn{1}{l|}{0}            & \multicolumn{1}{l|}{1}            & \multicolumn{1}{l|}{0}            & \multicolumn{1}{l|}{0.85}           & \multicolumn{1}{l|}{0.78}          & \multicolumn{1}{l|}{0.8}          & \multicolumn{1}{l|}{0.8}            & \multicolumn{1}{l|}{0.78}           & \multicolumn{1}{l|}{\textbf{0.802}} & \multicolumn{1}{l|}{7}                & \multicolumn{1}{l|}{14}               \\ \hline
\multicolumn{1}{|l|}{1}          & \multicolumn{1}{l|}{1}          & \multicolumn{1}{l|}{1}                             & \multicolumn{1}{l|}{1}          & \multicolumn{1}{l|}{1}                             & \multicolumn{1}{l|}{1}            & \multicolumn{1}{l|}{1}            & \multicolumn{1}{l|}{0}            & \multicolumn{1}{l|}{0}            & \multicolumn{1}{l|}{0.84}           & \multicolumn{1}{l|}{0.78}          & \multicolumn{1}{l|}{0.8}          & \multicolumn{1}{l|}{0.8}            & \multicolumn{1}{l|}{0.78}           & \multicolumn{1}{l|}{\textbf{0.8}}   & \multicolumn{1}{l|}{7}                & \multicolumn{1}{l|}{15}               \\ \hline
\multicolumn{1}{|l|}{0}          & \multicolumn{1}{l|}{1}          & \multicolumn{1}{l|}{1}                             & \multicolumn{1}{l|}{1}          & \multicolumn{1}{l|}{1}                             & \multicolumn{1}{l|}{1}            & \multicolumn{1}{l|}{1}            & \multicolumn{1}{l|}{0}            & \multicolumn{1}{l|}{1}            & \multicolumn{1}{l|}{0.84}           & \multicolumn{1}{l|}{0.79}          & \multicolumn{1}{l|}{0.78}         & \multicolumn{1}{l|}{0.8}            & \multicolumn{1}{l|}{0.78}           & \multicolumn{1}{l|}{\textbf{0.798}} & \multicolumn{1}{l|}{7}                & \multicolumn{1}{l|}{16*}              \\ \hline
                                 \multicolumn{17}{c}{\bf 6 combinations}                                          \\ \hline
\multicolumn{1}{|l|}{1}          & \multicolumn{1}{l|}{1}          & \multicolumn{1}{l|}{1}                             & \multicolumn{1}{l|}{1}          & \multicolumn{1}{l|}{0}                             & \multicolumn{1}{l|}{0}            & \multicolumn{1}{l|}{0}            & \multicolumn{1}{l|}{1}            & \multicolumn{1}{l|}{1}            & \multicolumn{1}{l|}{0.85}           & \multicolumn{1}{l|}{0.78}          & \multicolumn{1}{l|}{0.81}         & \multicolumn{1}{l|}{0.81}           & \multicolumn{1}{l|}{0.78}           & \multicolumn{1}{l|}{\textbf{0.806}} & \multicolumn{1}{l|}{6}                & \multicolumn{1}{l|}{17}               \\ \hline
\multicolumn{1}{|l|}{0}          & \multicolumn{1}{l|}{1}          & \multicolumn{1}{l|}{1}                             & \multicolumn{1}{l|}{0}          & \multicolumn{1}{l|}{1}                             & \multicolumn{1}{l|}{1}            & \multicolumn{1}{l|}{1}            & \multicolumn{1}{l|}{0}            & \multicolumn{1}{l|}{1}            & \multicolumn{1}{l|}{0.84}           & \multicolumn{1}{l|}{0.79}          & \multicolumn{1}{l|}{0.78}         & \multicolumn{1}{l|}{0.8}            & \multicolumn{1}{l|}{0.78}           & \multicolumn{1}{l|}{\textbf{0.798}} & \multicolumn{1}{l|}{6}                & \multicolumn{1}{l|}{18*}              \\ \hline
                                 \multicolumn{17}{c}{\bf 5 combinations}                                         \\ \hline
\multicolumn{1}{|l|}{1}          & \multicolumn{1}{l|}{0}          & \multicolumn{1}{l|}{1}                             & \multicolumn{1}{l|}{1}          & \multicolumn{1}{l|}{0}                             & \multicolumn{1}{l|}{0}            & \multicolumn{1}{l|}{0}            & \multicolumn{1}{l|}{1}            & \multicolumn{1}{l|}{1}            & \multicolumn{1}{l|}{0.84}           & \multicolumn{1}{l|}{0.78}          & \multicolumn{1}{l|}{0.78}         & \multicolumn{1}{l|}{0.79}           & \multicolumn{1}{l|}{0.77}           & \multicolumn{1}{l|}{\textbf{0.792}} & \multicolumn{1}{l|}{5}                & \multicolumn{1}{l|}{19}               \\ \hline
\multicolumn{1}{|l|}{1}          & \multicolumn{1}{l|}{1}          & \multicolumn{1}{l|}{0}                             & \multicolumn{1}{l|}{1}          & \multicolumn{1}{l|}{0}                             & \multicolumn{1}{l|}{0}            & \multicolumn{1}{l|}{0}            & \multicolumn{1}{l|}{1}            & \multicolumn{1}{l|}{1}            & \multicolumn{1}{l|}{0.84}           & \multicolumn{1}{l|}{0.78}          & \multicolumn{1}{l|}{0.78}         & \multicolumn{1}{l|}{0.79}           & \multicolumn{1}{l|}{0.77}           & \multicolumn{1}{l|}{\textbf{0.792}} & \multicolumn{1}{l|}{5}                & \multicolumn{1}{l|}{20}               \\ \hline
\multicolumn{1}{|l|}{1}          & \multicolumn{1}{l|}{1}          & \multicolumn{1}{l|}{1}                             & \multicolumn{1}{l|}{0}          & \multicolumn{1}{l|}{0}                             & \multicolumn{1}{l|}{0}            & \multicolumn{1}{l|}{0}            & \multicolumn{1}{l|}{1}            & \multicolumn{1}{l|}{1}            & \multicolumn{1}{l|}{0.84}           & \multicolumn{1}{l|}{0.78}          & \multicolumn{1}{l|}{0.8}          & \multicolumn{1}{l|}{0.8}            & \multicolumn{1}{l|}{0.78}           & \multicolumn{1}{l|}{\textbf{0.8}}   & \multicolumn{1}{l|}{5}                & \multicolumn{1}{l|}{21}               \\ \hline
\multicolumn{1}{|l|}{1}          & \multicolumn{1}{l|}{1}          & \multicolumn{1}{l|}{1}                             & \multicolumn{1}{l|}{1}          & \multicolumn{1}{l|}{0}                             & \multicolumn{1}{l|}{0}            & \multicolumn{1}{l|}{0}            & \multicolumn{1}{l|}{0}            & \multicolumn{1}{l|}{1}            & \multicolumn{1}{l|}{0.84}           & \multicolumn{1}{l|}{0.78}          & \multicolumn{1}{l|}{0.8}          & \multicolumn{1}{l|}{0.8}            & \multicolumn{1}{l|}{0.78}           & \multicolumn{1}{l|}{\textbf{0.8}}   & \multicolumn{1}{l|}{5}                & \multicolumn{1}{l|}{22}               \\ \hline
\multicolumn{1}{|l|}{1}          & \multicolumn{1}{l|}{1}          & \multicolumn{1}{l|}{1}                             & \multicolumn{1}{l|}{1}          & \multicolumn{1}{l|}{0}                             & \multicolumn{1}{l|}{0}            & \multicolumn{1}{l|}{0}            & \multicolumn{1}{l|}{1}            & \multicolumn{1}{l|}{0}            & \multicolumn{1}{l|}{0.84}           & \multicolumn{1}{l|}{0.78}          & \multicolumn{1}{l|}{0.8}          & \multicolumn{1}{l|}{0.8}            & \multicolumn{1}{l|}{0.78}           & \multicolumn{1}{l|}{\textbf{0.8}}   & \multicolumn{1}{l|}{5}                & \multicolumn{1}{l|}{23}               \\ \hline
\multicolumn{1}{|l|}{0}          & \multicolumn{1}{l|}{1}          & \multicolumn{1}{l|}{1}                             & \multicolumn{1}{l|}{0}          & \multicolumn{1}{l|}{0}                             & \multicolumn{1}{l|}{1}            & \multicolumn{1}{l|}{1}            & \multicolumn{1}{l|}{0}            & \multicolumn{1}{l|}{1}            & \multicolumn{1}{l|}{0.83}           & \multicolumn{1}{l|}{0.79}          & \multicolumn{1}{l|}{0.78}         & \multicolumn{1}{l|}{0.8}            & \multicolumn{1}{l|}{0.78}           & \multicolumn{1}{l|}{\textbf{0.796}} & \multicolumn{1}{l|}{5}                & \multicolumn{1}{l|}{24*}              \\ \hline
                                 \multicolumn{17}{c}{\bf 4 combinations}                                          \\ \hline
\multicolumn{1}{|l|}{1}          & \multicolumn{1}{l|}{1}          & \multicolumn{1}{l|}{1}                             & \multicolumn{1}{l|}{1}          & \multicolumn{1}{l|}{0}                             & \multicolumn{1}{l|}{0}            & \multicolumn{1}{l|}{0}            & \multicolumn{1}{l|}{0}            & \multicolumn{1}{l|}{0}            & \multicolumn{1}{l|}{0.83}           & \multicolumn{1}{l|}{0.78}          & \multicolumn{1}{l|}{0.77}         & \multicolumn{1}{l|}{0.78}           & \multicolumn{1}{l|}{0.77}           & \multicolumn{1}{l|}{\textbf{0.786}} & \multicolumn{1}{l|}{4}                & \multicolumn{1}{l|}{25}               \\ \hline
\multicolumn{1}{|l|}{1}          & \multicolumn{1}{l|}{1}          & \multicolumn{1}{l|}{1}                             & \multicolumn{1}{l|}{0}          & \multicolumn{1}{l|}{0}                             & \multicolumn{1}{l|}{0}            & \multicolumn{1}{l|}{0}            & \multicolumn{1}{l|}{1}            & \multicolumn{1}{l|}{0}            & \multicolumn{1}{l|}{0.83}           & \multicolumn{1}{l|}{0.78}          & \multicolumn{1}{l|}{0.78}         & \multicolumn{1}{l|}{0.78}           & \multicolumn{1}{l|}{0.77}           & \multicolumn{1}{l|}{\textbf{0.788}} & \multicolumn{1}{l|}{4}                & \multicolumn{1}{l|}{26}               \\ \hline
\multicolumn{1}{|l|}{1}          & \multicolumn{1}{l|}{1}          & \multicolumn{1}{l|}{1}                             & \multicolumn{1}{l|}{0}          & \multicolumn{1}{l|}{0}                             & \multicolumn{1}{l|}{0}            & \multicolumn{1}{l|}{0}            & \multicolumn{1}{l|}{0}            & \multicolumn{1}{l|}{1}            & \multicolumn{1}{l|}{0.83}           & \multicolumn{1}{l|}{0.78}          & \multicolumn{1}{l|}{0.77}         & \multicolumn{1}{l|}{0.78}           & \multicolumn{1}{l|}{0.77}           & \multicolumn{1}{l|}{\textbf{0.786}} & \multicolumn{1}{l|}{4}                & \multicolumn{1}{l|}{27}               \\ \hline
\multicolumn{1}{|l|}{1}          & \multicolumn{1}{l|}{1}          & \multicolumn{1}{l|}{0}                             & \multicolumn{1}{l|}{0}          & \multicolumn{1}{l|}{0}                             & \multicolumn{1}{l|}{0}            & \multicolumn{1}{l|}{0}            & \multicolumn{1}{l|}{1}            & \multicolumn{1}{l|}{1}            & \multicolumn{1}{l|}{0.8}            & \multicolumn{1}{l|}{0.78}          & \multicolumn{1}{l|}{0.72}         & \multicolumn{1}{l|}{0.74}           & \multicolumn{1}{l|}{0.76}           & \multicolumn{1}{l|}{\textbf{0.76}}  & \multicolumn{1}{l|}{4}                & \multicolumn{1}{l|}{28}               \\ \hline
\multicolumn{1}{|l|}{1}          & \multicolumn{1}{l|}{0}          & \multicolumn{1}{l|}{1}                             & \multicolumn{1}{l|}{0}          & \multicolumn{1}{l|}{0}                             & \multicolumn{1}{l|}{0}            & \multicolumn{1}{l|}{0}            & \multicolumn{1}{l|}{1}            & \multicolumn{1}{l|}{1}            & \multicolumn{1}{l|}{0.79}           & \multicolumn{1}{l|}{0.77}          & \multicolumn{1}{l|}{0.7}          & \multicolumn{1}{l|}{0.73}           & \multicolumn{1}{l|}{0.75}           & \multicolumn{1}{l|}{\textbf{0.748}} & \multicolumn{1}{l|}{4}                & \multicolumn{1}{l|}{29}               \\ \hline
\multicolumn{1}{|l|}{0}          & \multicolumn{1}{l|}{1}          & \multicolumn{1}{l|}{1}                             & \multicolumn{1}{l|}{1}          & \multicolumn{1}{l|}{0}                             & \multicolumn{1}{l|}{0}            & \multicolumn{1}{l|}{0}            & \multicolumn{1}{l|}{1}            & \multicolumn{1}{l|}{0}            & \multicolumn{1}{l|}{0.76}           & \multicolumn{1}{l|}{0.65}          & \multicolumn{1}{l|}{0.75}         & \multicolumn{1}{l|}{0.76}           & \multicolumn{1}{l|}{0.74}           & \multicolumn{1}{l|}{\textbf{0.732}} & \multicolumn{1}{l|}{4}                & \multicolumn{1}{l|}{30}               \\ \hline
\multicolumn{1}{|l|}{1}          & \multicolumn{1}{l|}{0}          & \multicolumn{1}{l|}{1}                             & \multicolumn{1}{l|}{1}          & \multicolumn{1}{l|}{0}                             & \multicolumn{1}{l|}{0}            & \multicolumn{1}{l|}{0}            & \multicolumn{1}{l|}{1}            & \multicolumn{1}{l|}{0}            & \multicolumn{1}{l|}{0.79}           & \multicolumn{1}{l|}{0.77}          & \multicolumn{1}{l|}{0.7}          & \multicolumn{1}{l|}{0.72}           & \multicolumn{1}{l|}{0.75}           & \multicolumn{1}{l|}{\textbf{0.746}} & \multicolumn{1}{l|}{4}                & \multicolumn{1}{l|}{31}               \\ \hline
\multicolumn{1}{|l|}{1}          & \multicolumn{1}{l|}{1}          & \multicolumn{1}{l|}{0}                             & \multicolumn{1}{l|}{1}          & \multicolumn{1}{l|}{0}                             & \multicolumn{1}{l|}{0}            & \multicolumn{1}{l|}{0}            & \multicolumn{1}{l|}{1}            & \multicolumn{1}{l|}{0}            & \multicolumn{1}{l|}{0.8}            & \multicolumn{1}{l|}{0.77}          & \multicolumn{1}{l|}{0.71}         & \multicolumn{1}{l|}{0.74}           & \multicolumn{1}{l|}{0.75}           & \multicolumn{1}{l|}{\textbf{0.754}} & \multicolumn{1}{l|}{4}                & \multicolumn{1}{l|}{32}               \\ \hline
\multicolumn{1}{|l|}{0}          & \multicolumn{1}{l|}{1}          & \multicolumn{1}{l|}{1}                             & \multicolumn{1}{l|}{0}          & \multicolumn{1}{l|}{0}                             & \multicolumn{1}{l|}{0}            & \multicolumn{1}{l|}{1}            & \multicolumn{1}{l|}{0}            & \multicolumn{1}{l|}{1}            & \multicolumn{1}{l|}{0.83}           & \multicolumn{1}{l|}{0.79}          & \multicolumn{1}{l|}{0.78}         & \multicolumn{1}{l|}{0.76}           & \multicolumn{1}{l|}{0.77}           & \multicolumn{1}{l|}{\textbf{0.786}} & \multicolumn{1}{l|}{4}                & \multicolumn{1}{l|}{33*}              \\ \hline
                                \multicolumn{17}{c}{\bf 3 combinations}                                          \\ \hline
\multicolumn{1}{|l|}{1}          & \multicolumn{1}{l|}{1}          & \multicolumn{1}{l|}{1}                             & \multicolumn{1}{l|}{0}          & \multicolumn{1}{l|}{0}                             & \multicolumn{1}{l|}{0}            & \multicolumn{1}{l|}{0}            & \multicolumn{1}{l|}{0}            & \multicolumn{1}{l|}{0}            & \multicolumn{1}{l|}{0.76}           & \multicolumn{1}{l|}{0.65}          & \multicolumn{1}{l|}{0.75}         & \multicolumn{1}{l|}{0.76}           & \multicolumn{1}{l|}{0.74}           & \multicolumn{1}{l|}{\textbf{0.732}} & \multicolumn{1}{l|}{3}                & \multicolumn{1}{l|}{34}               \\ \hline
\multicolumn{1}{|l|}{1}          & \multicolumn{1}{l|}{1}          & \multicolumn{1}{l|}{0}                             & \multicolumn{1}{l|}{0}          & \multicolumn{1}{l|}{0}                             & \multicolumn{1}{l|}{0}            & \multicolumn{1}{l|}{0}            & \multicolumn{1}{l|}{1}            & \multicolumn{1}{l|}{0}            & \multicolumn{1}{l|}{0.79}           & \multicolumn{1}{l|}{0.77}          & \multicolumn{1}{l|}{0.69}         & \multicolumn{1}{l|}{0.72}           & \multicolumn{1}{l|}{0.75}           & \multicolumn{1}{l|}{\textbf{0.744}} & \multicolumn{1}{l|}{3}                & \multicolumn{1}{l|}{35}               \\ \hline
\multicolumn{1}{|l|}{1}          & \multicolumn{1}{l|}{1}          & \multicolumn{1}{l|}{0}                             & \multicolumn{1}{l|}{0}          & \multicolumn{1}{l|}{0}                             & \multicolumn{1}{l|}{0}            & \multicolumn{1}{l|}{0}            & \multicolumn{1}{l|}{0}            & \multicolumn{1}{l|}{1}            & \multicolumn{1}{l|}{0.78}           & \multicolumn{1}{l|}{0.77}          & \multicolumn{1}{l|}{0.68}         & \multicolumn{1}{l|}{0.71}           & \multicolumn{1}{l|}{0.75}           & \multicolumn{1}{l|}{\textbf{0.738}} & \multicolumn{1}{l|}{3}                & \multicolumn{1}{l|}{36}               \\ \hline
\multicolumn{1}{|l|}{1}          & \multicolumn{1}{l|}{1}          & \multicolumn{1}{l|}{0}                             & \multicolumn{1}{l|}{1}          & \multicolumn{1}{l|}{0}                             & \multicolumn{1}{l|}{0}            & \multicolumn{1}{l|}{0}            & \multicolumn{1}{l|}{0}            & \multicolumn{1}{l|}{0}            & \multicolumn{1}{l|}{0.79}           & \multicolumn{1}{l|}{0.77}          & \multicolumn{1}{l|}{0.69}         & \multicolumn{1}{l|}{0.71}           & \multicolumn{1}{l|}{0.75}           & \multicolumn{1}{l|}{\textbf{0.742}} & \multicolumn{1}{l|}{3}                & \multicolumn{1}{l|}{37}               \\ \hline
\multicolumn{1}{|l|}{1}          & \multicolumn{1}{l|}{0}          & \multicolumn{1}{l|}{1}                             & \multicolumn{1}{l|}{0}          & \multicolumn{1}{l|}{0}                             & \multicolumn{1}{l|}{0}            & \multicolumn{1}{l|}{0}            & \multicolumn{1}{l|}{0}            & \multicolumn{1}{l|}{1}            & \multicolumn{1}{l|}{0.78}           & \multicolumn{1}{l|}{0.77}          & \multicolumn{1}{l|}{0.68}         & \multicolumn{1}{l|}{0.7}            & \multicolumn{1}{l|}{0.74}           & \multicolumn{1}{l|}{\textbf{0.734}} & \multicolumn{1}{l|}{3}                & \multicolumn{1}{l|}{38}               \\ \hline
\multicolumn{1}{|l|}{1}          & \multicolumn{1}{l|}{0}          & \multicolumn{1}{l|}{1}                             & \multicolumn{1}{l|}{0}          & \multicolumn{1}{l|}{0}                             & \multicolumn{1}{l|}{0}            & \multicolumn{1}{l|}{0}            & \multicolumn{1}{l|}{1}            & \multicolumn{1}{l|}{0}            & \multicolumn{1}{l|}{0.78}           & \multicolumn{1}{l|}{0.77}          & \multicolumn{1}{l|}{0.68}         & \multicolumn{1}{l|}{0.71}           & \multicolumn{1}{l|}{0.74}           & \multicolumn{1}{l|}{\textbf{0.736}} & \multicolumn{1}{l|}{3}                & \multicolumn{1}{l|}{39}               \\ \hline
\multicolumn{1}{|l|}{1}          & \multicolumn{1}{l|}{0}          & \multicolumn{1}{l|}{1}                             & \multicolumn{1}{l|}{1}          & \multicolumn{1}{l|}{0}                             & \multicolumn{1}{l|}{0}            & \multicolumn{1}{l|}{0}            & \multicolumn{1}{l|}{0}            & \multicolumn{1}{l|}{0}            & \multicolumn{1}{l|}{0.78}           & \multicolumn{1}{l|}{0.77}          & \multicolumn{1}{l|}{0.68}         & \multicolumn{1}{l|}{0.71}           & \multicolumn{1}{l|}{0.74}           & \multicolumn{1}{l|}{\textbf{0.736}} & \multicolumn{1}{l|}{3}                & \multicolumn{1}{l|}{40}               \\ \hline
\multicolumn{1}{|l|}{1}          & \multicolumn{1}{l|}{0}          & \multicolumn{1}{l|}{0}                             & \multicolumn{1}{l|}{0}          & \multicolumn{1}{l|}{0}                             & \multicolumn{1}{l|}{0}            & \multicolumn{1}{l|}{0}            & \multicolumn{1}{l|}{1}            & \multicolumn{1}{l|}{1}            & \multicolumn{1}{l|}{0.73}           & \multicolumn{1}{l|}{0.75}          & \multicolumn{1}{l|}{0.62}         & \multicolumn{1}{l|}{0.65}           & \multicolumn{1}{l|}{0.7}            & \multicolumn{1}{l|}{\textbf{0.69}}  & \multicolumn{1}{l|}{3}                & \multicolumn{1}{l|}{41}               \\ \hline
\multicolumn{1}{|l|}{1}          & \multicolumn{1}{l|}{0}          & \multicolumn{1}{l|}{0}                             & \multicolumn{1}{l|}{1}          & \multicolumn{1}{l|}{0}                             & \multicolumn{1}{l|}{0}            & \multicolumn{1}{l|}{0}            & \multicolumn{1}{l|}{1}            & \multicolumn{1}{l|}{0}            & \multicolumn{1}{l|}{0.74}           & \multicolumn{1}{l|}{0.75}          & \multicolumn{1}{l|}{0.62}         & \multicolumn{1}{l|}{0.65}           & \multicolumn{1}{l|}{0.7}            & \multicolumn{1}{l|}{\textbf{0.692}} & \multicolumn{1}{l|}{3}                & \multicolumn{1}{l|}{42}               \\ \hline
\multicolumn{1}{|l|}{1}          & \multicolumn{1}{l|}{0}          & \multicolumn{1}{l|}{0}                             & \multicolumn{1}{l|}{1}          & \multicolumn{1}{l|}{0}                             & \multicolumn{1}{l|}{0}            & \multicolumn{1}{l|}{0}            & \multicolumn{1}{l|}{0}            & \multicolumn{1}{l|}{1}            & \multicolumn{1}{l|}{0.73}           & \multicolumn{1}{l|}{0.75}          & \multicolumn{1}{l|}{0.61}         & \multicolumn{1}{l|}{0.64}           & \multicolumn{1}{l|}{0.7}            & \multicolumn{1}{l|}{\textbf{0.686}} & \multicolumn{1}{l|}{3}                & \multicolumn{1}{l|}{43}               \\ \hline
\multicolumn{1}{|l|}{0}          & \multicolumn{1}{l|}{0}          & \multicolumn{1}{l|}{1}                             & \multicolumn{1}{l|}{0}          & \multicolumn{1}{l|}{0}                             & \multicolumn{1}{l|}{0}            & \multicolumn{1}{l|}{1}            & \multicolumn{1}{l|}{0}            & \multicolumn{1}{l|}{1}            & \multicolumn{1}{l|}{0.8}            & \multicolumn{1}{l|}{0.78}          & \multicolumn{1}{l|}{0.73}         & \multicolumn{1}{l|}{0.72}           & \multicolumn{1}{l|}{0.75}           & \multicolumn{1}{l|}{\textbf{0.756}} & \multicolumn{1}{l|}{3}                & \multicolumn{1}{l|}{44*}              \\ \hline
                                \multicolumn{17}{c}{\bf 2 combinations}                                                                                 \\ \hline
\multicolumn{1}{|l|}{1}          & \multicolumn{1}{l|}{1}          & \multicolumn{1}{l|}{0}                             & \multicolumn{1}{l|}{0}          & \multicolumn{1}{l|}{0}                             & \multicolumn{1}{l|}{0}            & \multicolumn{1}{l|}{0}            & \multicolumn{1}{l|}{0}            & \multicolumn{1}{l|}{0}            & \multicolumn{1}{l|}{0.7}            & \multicolumn{1}{l|}{0.63}          & \multicolumn{1}{l|}{0.65}         & \multicolumn{1}{l|}{0.68}           & \multicolumn{1}{l|}{0.72}           & \multicolumn{1}{l|}{\textbf{0.676}} & \multicolumn{1}{l|}{2}                & \multicolumn{1}{l|}{45}               \\ \hline
\multicolumn{1}{|l|}{1}          & \multicolumn{1}{l|}{0}          & \multicolumn{1}{l|}{1}                             & \multicolumn{1}{l|}{0}          & \multicolumn{1}{l|}{0}                             & \multicolumn{1}{l|}{0}            & \multicolumn{1}{l|}{0}            & \multicolumn{1}{l|}{0}            & \multicolumn{1}{l|}{0}            & \multicolumn{1}{l|}{0.69}           & \multicolumn{1}{l|}{0.63}          & \multicolumn{1}{l|}{0.64}         & \multicolumn{1}{l|}{0.68}           & \multicolumn{1}{l|}{0.71}           & \multicolumn{1}{l|}{\textbf{0.67}}  & \multicolumn{1}{l|}{2}                & \multicolumn{1}{l|}{46}               \\ \hline
\multicolumn{1}{|l|}{1}          & \multicolumn{1}{l|}{0}          & \multicolumn{1}{l|}{0}                             & \multicolumn{1}{l|}{1}          & \multicolumn{1}{l|}{0}                             & \multicolumn{1}{l|}{0}            & \multicolumn{1}{l|}{0}            & \multicolumn{1}{l|}{0}            & \multicolumn{1}{l|}{0}            & \multicolumn{1}{l|}{0.43}           & \multicolumn{1}{l|}{0.52}          & \multicolumn{1}{l|}{0.29}         & \multicolumn{1}{l|}{0.28}           & \multicolumn{1}{l|}{0.3}            & \multicolumn{1}{l|}{\textbf{0.364}} & \multicolumn{1}{l|}{2}                & \multicolumn{1}{l|}{47}               \\ \hline
\multicolumn{1}{|l|}{1}          & \multicolumn{1}{l|}{0}          & \multicolumn{1}{l|}{0}                             & \multicolumn{1}{l|}{0}          & \multicolumn{1}{l|}{0}                             & \multicolumn{1}{l|}{0}            & \multicolumn{1}{l|}{0}            & \multicolumn{1}{l|}{1}            & \multicolumn{1}{l|}{0}            & \multicolumn{1}{l|}{0.34}           & \multicolumn{1}{l|}{0.43}          & \multicolumn{1}{l|}{0.21}         & \multicolumn{1}{l|}{0.2}            & \multicolumn{1}{l|}{0.2}            & \multicolumn{1}{l|}{\textbf{0.276}} & \multicolumn{1}{l|}{2}                & \multicolumn{1}{l|}{48}               \\ \hline
\multicolumn{1}{|l|}{1}          & \multicolumn{1}{l|}{0}          & \multicolumn{1}{l|}{0}                             & \multicolumn{1}{l|}{0}          & \multicolumn{1}{l|}{0}                             & \multicolumn{1}{l|}{0}            & \multicolumn{1}{l|}{0}            & \multicolumn{1}{l|}{0}            & \multicolumn{1}{l|}{1}            & \multicolumn{1}{l|}{0.51}           & \multicolumn{1}{l|}{0.59}          & \multicolumn{1}{l|}{0.36}         & \multicolumn{1}{l|}{0.36}           & \multicolumn{1}{l|}{0.38}           & \multicolumn{1}{l|}{\textbf{0.44}}  & \multicolumn{1}{l|}{2}                & \multicolumn{1}{l|}{49}               \\ \hline
\multicolumn{1}{|l|}{0}          & \multicolumn{1}{l|}{0}          & \multicolumn{1}{l|}{1}                             & \multicolumn{1}{l|}{0}          & \multicolumn{1}{l|}{0}                             & \multicolumn{1}{l|}{0}            & \multicolumn{1}{l|}{1}            & \multicolumn{1}{l|}{0}            & \multicolumn{1}{l|}{0}            & \multicolumn{1}{l|}{0.71}           & \multicolumn{1}{l|}{0.76}          & \multicolumn{1}{l|}{0.68}         & \multicolumn{1}{l|}{0.71}           & \multicolumn{1}{l|}{0.74}           & \multicolumn{1}{l|}{\textbf{0.72}}  & \multicolumn{1}{l|}{2}                & \multicolumn{1}{l|}{50*}              \\ \hline

%\multicolumn{17}{c}{\bf 1 combination}                                       \\ \hline
%\multicolumn{1}{|l|}{0}          & \multicolumn{1}{l|}{0}          & \multicolumn{1}{l|}{1}                             & \multicolumn{1}{l|}{0}          & \multicolumn{1}{l|}{0}                             & \multicolumn{1}{l|}{0}            & \multicolumn{1}{l|}{0}            & \multicolumn{1}{l|}{0}            & \multicolumn{1}{l|}{0}            & \multicolumn{1}{l|}{0.7}            & \multicolumn{1}{l|}{0.64}          & \multicolumn{1}{l|}{0.68}         & \multicolumn{1}{l|}{0.68}           & \multicolumn{1}{l|}{0.72}           & \multicolumn{1}{l|}{\textbf{0.684}} & \multicolumn{1}{l|}{1}                & \multicolumn{1}{l|}{51*}              \\ \hline
\end{tabular}
\end{table*}

% \clearpage

% \newpage

\section{Coefficients of the functional forms}
\label{sec:app_coefficients_functionalform}

\blue{In Sec.~4 we propose Eqs.~\ref{eq:1_factor} and~\ref{eq:2_factors} as estimators of the perceptual sensitivity for 1- and 2-factor models respectively. In Tables~\ref{tab:1D_coefs} and~\ref{tab:2D_coefs} we give the actual weights obtained in the experiments for each distance.}

\begin{table*}[]
\setlength\tabcolsep{4pt}
\centering
\caption{Coefficients for the Eq.~3 for each metric, and each weight normalised by $w_0$ in order to compare between metrics.}
\label{tab:1D_coefs}
\begin{tabular}{lllllll}
Coefs             & \textbf{MSSIM}       & \textbf{NLPD}        & \textbf{PIM}          & \textbf{LPIPS}          & \textbf{DISTS}  & \blue{\textbf{Mean}}   \\ \hline
$w_0$             & 29.5                 & 65                   & 15400                 & 198                     & 161                   \\
$w_1$             & $4.9\times 10^{-3}$  & $9.5\times 10^{-3}$  & 2.62                  & $3.33\times 10^{-2}$    & $2.58\times 10^{-2}$  \\
$w_2$             & $2.05\times 10^{-7}$ & $3.62\times 10^{-7}$ & $1.11\times 10^{-4}$  & $1.41 \times10^{-6}$    & $1.05\times 10^{-6}$  \\ \hline
$w_0/w_0$ & 1                    & 1                    & 1                     & 1                       & 1       & 1              \\
$w_1/w_0$ & $1.66\times 10^{-4}$ & $1.46\times 10^{-4}$ & $1.70 \times 10^{-4}$ & $1.68 \times 10^{-4}$ & $1.60 \times 10^{-4}$ & $1.62\times 10^{-4}$ \\
$w_2/w_0$ & $6.94\times 10^{-9}$ & $5.57\times 10^{-9}$ & $7.21\times 10^{-9}$  & $7.12 \times 10^{-9}$   & $6.52-\times 10^{-9}$ & $6.67\times 10^{-9}$ 
\end{tabular}
\end{table*}

\begin{table*}[]
\setlength\tabcolsep{4pt}
\centering
\caption{Coefficients for the Eq.~4 for each metric, and each weight normalised by $w_0$ in order to compare between metrics.}
\label{tab:2D_coefs}
\begin{tabular}{lllllll}
Coefs     & \textbf{MSSIM}       & \textbf{NLPD}        & \textbf{PIM}         & \textbf{LPIPS}       & \textbf{DISTS} & \blue{\textbf{Mean}}      \\ \hline
$w_0$     & $28$                 & $58$                 & $15100$              & $194$                & $156$                \\
$w_1$     & $4.69\times 10^{-3}$ & $8.19\times 10^{-3}$ & $2.57$               & $3.26\times 10^{-2}$ & $2.49\times 10^{-2}$ \\
$w_2$     & $1.96\times 10^{-7}$ & $3.09\times 10^{-7}$ & $1.09\times 10^{-4}$ & $1.37\times 10^{-6}$ & $1.00\times 10^{-6}$ \\
$w_3$     & $-0.597$             & $-3.74$              & $-141$               & $-1.93$              & $-2.54$              \\ \hline
$w_0/w_0$ & $1$                  & $1$                  & $1$                  & $1$                  & $1$  & $1$                \\
$w_1/w_0$ & $1.68\times 10^{-4}$ & $1.41\times10^{-4}$  & $1.70\times 10^{-4}$ & $1.68\times10^{-4}$  & $1.60\times 10^{-4}$ & $1.61\times 10^{-4}$ \\
$w_2/w_0$ & $7.00\times 10^{-9}$ & $5.33\times 10^{-9}$ & $7.22\times 10^{-9}$ & $7.06\times 10^{-9}$ & $6.41\times 10^{-9}$ & $6.6\times 10^{-9}$  \\
$w_3/w_0$ & $-0.021$             & $-0.064$             & $-0.0093$            & $-0.0010$            & $-0.016$ & $-0.024$           
\end{tabular}
\end{table*}

\blue{Each metric has a different interpretation of the sensitivity units, therefore the weights (coefficients) of the proposed equations are different for each measure. However, note that the proportion of the coefficients for each probability factor are very similar. Actually by normalizing the coefficients by the \emph{bias} ($b$) the normalized coefficients are very similar (see Tables~\ref{tab:1D_coefs} and~\ref{tab:2D_coefs}).} 

\blue{By simply taking the mean of each normalized coefficient for the six metrics we get the last column which is a set of coefficients that can be use to predict the sensitivity for each metric (up to the normalizing factor which does not affecto to the correlatiobn). The Pearson correlations using the models with the \emph{Mean} coefficients are shown in Table~\ref{tab:correlations_norm_model}. These simple and general models get very good correlation with the sensitivity of all the metrics simultaneously.}

\blue{As a summary, if one takes one of the models in eqs. \ref{eq:1_factor} or \ref{eq:1_factor}, using the coefficients in the mean column in Tables~\ref{tab:1D_coefs} or \ref{tab:2D_coefs} respectively, can get a good estimation of the sensitivity of a perceptual measure.}

\begin{table*}[h]
\centering
\caption{\blue{Pearson correlations obtained using a single set of coefficients ("Mean" column in Tables~\ref{tab:1D_coefs} and \ref{tab:2D_coefs}) for the 1F and 2F models. Pearson for coefficients fitted for each particular metric are given for comparison (correlations are the same as in Table~\ref{tab:1D} row $d=2$) for 1F, and Table~\ref{tab:app_2D} first row 4 combinations.}}
\label{tab:correlations_norm_model}
\begin{tabular}{llllll}
     & \textbf{MSSIM}       & \textbf{NLPD}        & \textbf{PIM}         & \textbf{LPIPS}       & \textbf{DISTS}       \\ \hline
Mean coefficients 1F     & $0.75$                 & $0.64$                 & $0.73$              & $0.75$                & $0.74$                \\
Specific coefficients 1F     & $0.76$                 & $0.65$                 & $0.75$              & $0.76$                & $0.74$   \\  \hline
Mean coefficients 2F     & $0.82$                 & $0.76$                 & $0.75$              & $0.76$                & $0.77$                \\
Specific coefficients 2F     & $0.83$                 & $0.78$                 & $0.77$              & $0.78$                & $0.77$   \\                 
\end{tabular}
\end{table*}

\section{PixelCNN++ smoothness}
\label{sec:smoothness}

\begin{figure*}[h]
    \centering
    \begin{tabular}{ccc}
        \includegraphics[width=0.3\columnwidth]{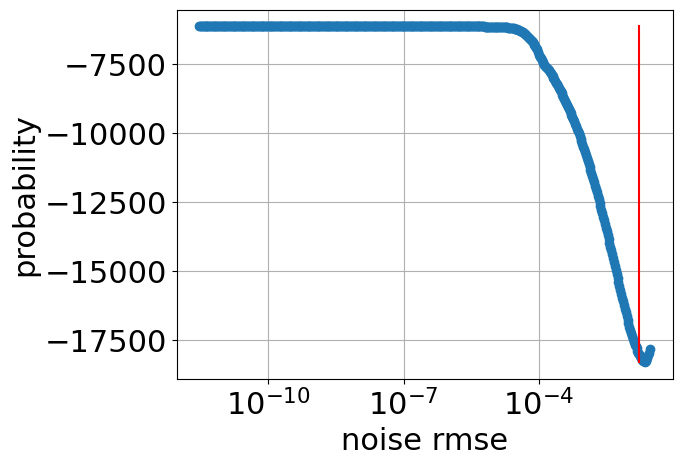} &
        \includegraphics[width=0.3\columnwidth]{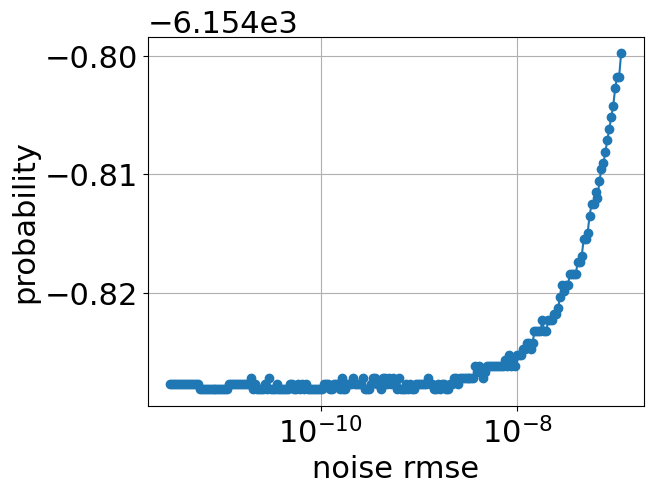} & 
        \includegraphics[width=0.3\columnwidth]{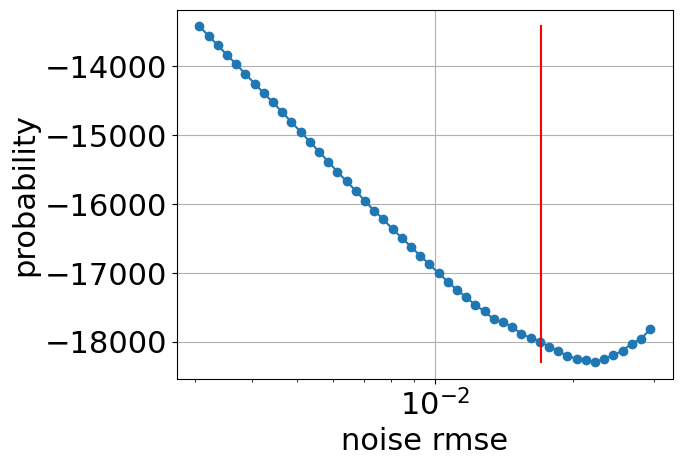} 
    \end{tabular}
    \caption{Probability estimations and the corresponding rmse between the original and the distorted image using uniform noise. We use a fixed image and the same noise pattern but we increase the magnitude of the noise, the red line marks the level of noise used in our experiments. Left: full plot, the red line is the noise we add throughout the paper. Middle: details at extremely low level of noise (quantisation effect is visible). Right: details at very low level of noise (quantisation effect is negligible).}
    \label{fig:quant}
\end{figure*}

\blue{While in the original PixelCNN++ paper authors talk about PDF as the magnitude estimated by the model \cite{salimans2017pixelcnn++}, there are some parts of the model that perform a kind of quantisation. one can wonder if this quantisation can affect the experiments. In Fig.~\ref{fig:quant} we show the probability of an image with an increasing level of standard deviation for the noise, we kept the noise pattern constant (i.e. fixed seed) but we modify the magnitude of the noise. While at extremely low levels the effect of the quantisation is noticeable, at very low and low level of noise the estimations vary softly. As a reference we show the level of noise analysed in this work, it is really far from the quantisation region.}

% \section{Perceptual tests on the model}
% \label{sec:app_perc}
% Here we show the stimuli used for both perceptual tests performed using the proposed models \blue{and the schematic for the perceptual experiments.}

% \begin{figure*}[h]
%     \centering
%     \begin{tabular}{c}
%      \includegraphics[width=0.8\textwidth]{figs/stimuli.JPG} 
%     \end{tabular}
%     \caption{Stimuli for the perceptual tests of the proposed model. (left) Gratings are generated at wavelengths that can be accurately shown in a $32\times32$ spatial size, generated at contrasts of $[0.2, 0.4, 0.7]$. (right) Shows one example of editing an image from the CIFAR10 dataset to vary contrast of $[0.2, 0.4, 0.7]$ and luminance values in $[60, 140]$.}
%     \label{fig:test}
% \end{figure*}

% \begin{figure}
%     \centering
%     \includegraphics[width=\textwidth]{figs/perceptual_schematic.png}
%     \caption{Schematic describing the perceptual experiments 1 and 2.}
%     \label{fig:perceptual_schematic}
% \end{figure}

\end{document}